\documentclass{article}

 \usepackage[main, final]{neurips_2026}
\usepackage[utf8]{inputenc} % allow utf-8 input
\ifdefined\DeclareUnicodeCharacter
  \DeclareUnicodeCharacter{2019}{\textquoteright}
  \DeclareUnicodeCharacter{2014}{---}
\fi
\usepackage[T1]{fontenc}    % use 8-bit T1 fonts
\usepackage{url}            % simple URL typesetting
\usepackage{booktabs}       % professional-quality tables
\usepackage{amsfonts}       % blackboard math symbols
\usepackage{nicefrac}       % compact symbols for 1/2, etc.
\usepackage{microtype}      % microtypography
\usepackage[table,svgnames]{xcolor}         % colors
\usepackage{graphicx}
\usepackage{multirow}
\usepackage{multicol}
\usepackage{algorithm}
\usepackage{algpseudocode}

\usepackage{amsmath}
\usepackage{amssymb}
\usepackage{amsthm}
\usepackage{mathtools}   % extends amsmath, fixes \coloneqq etc.

\usepackage{wrapfig}

\usepackage{nicematrix}

\usepackage{comment}
\usepackage[colorlinks=true,
            breaklinks=true,
            linkcolor=DarkOrchid, 
            urlcolor=cyan, 
            citecolor=Maroon]{hyperref}
\usepackage{cleveref}    % \cref{eq:diag-obj} → "Eq. (3)" automatically

\definecolor{lightpurple}{RGB}{232, 232, 232}

\usepackage{pifont}

\title{ActQuant: Sub-4-bit Action-Guided Backbone\\Weight-Only Quantization for Vision-Language-Action Models}

\newcommand{\diff}[1]{%
  \ifdim#1pt<0pt
    \textcolor{green!80!black}{#1}%
  \else
    \ifdim#1pt>0pt
      \textcolor{green!60!black}{+#1}%
    \else
      \textcolor{gray}{0.00}%
    \fi
  \fi
}

\usepackage{authblk}

\author[1*]{Arash Akbari}
\author[1]{Arman Akbari}
\author[1]{Masih Eskandar}
\author[2]{Qitao Tan}
\author[1,4]{Yixiao Chen}
\author[1]{Jingwu Luo}
\author[1]{Bertha Pangaribuan}
\author[1]{Liyun Zhang}
\author[3]{Gaowen Liu}
\author[1]{Xue Lin}
\author[1]{Jennifer Dy}
\author[2]{Geng Yuan}
\author[1]{\mbox{Stratis Ioannidis}}
\author[1]{Yanzhi Wang}

\affil[ ]{\small $^{1}$Northeastern University \quad $^{2}$University of Georgia \quad $^{3}$Cisco Systems \quad $^{4}$EmbodyX}

\begin{document}
\renewcommand{\thefootnote}{\fnsymbol{footnote}}
\footnotetext[1]{Corresponding author: \texttt{akbari.ara@northeastern.edu}}
\renewcommand{\thefootnote}{\arabic{footnote}}

\maketitle

\begin{center}
\vspace{-2em}
Project page: \url{https://actquant.github.io}
\end{center}

\begin{abstract}
Vision-Language-Action (VLA) models exhibit remarkable action generation for embodied intelligence, but their heavy compute makes deployment on edge platforms impractical. Aggressive, sub-4-bit weight quantization is the natural solution, yet existing post-training quantization (PTQ) methods suffer severe performance degradation in this regime. To address this, we introduce \textbf{ActQuant}, an action-guided mixed-precision PTQ framework for \emph{backbone weight-only} quantization: it quantizes the weights of the vision encoder and the language backbone, while the multimodal projector and the action head are kept at higher precision, so every bits-per-weight figure we report is a backbone budget and the full-policy average is higher. ActQuant operates in two stages: (1) an inter-tensor bit allocator that assigns each weight matrix a single bit-width based on how much it contributes to predicting the agent's actions; (2) an intra-tensor scale optimizer tunes per-block quantization scales using action-aware curvature, so that dynamic range is concentrated on the weights most influential for control. To deliver the on-device benefits of our aggressive quantization, we further introduce \textbf{OmniModel.cpp}, an agentic conversion pipeline that ports architectures into a native C/C++ runtime with efficient low-bit kernels.
We evaluate ActQuant in simulation and on real-world UR3 hardware, both single-arm and bimanual, with all models deployed through OmniModel.cpp. On LIBERO it retains 95.0\% (OpenVLA-OFT) and 94.8\% ($\pi_{0.5}$) at 3.0 bits-per-weight; at 2.5 bpw on OpenVLA-OFT it reaches 90.1\% while QVLA scores 0.0\%, compressing the backbone from 14.3 GB to 2.7 GB (\textbf{5.3$\times$}), and at 2.0 bpw on $\pi_{0.5}$ it reaches 48.0\% where all evaluated baselines score 0.0\%. On the UR3 arm it matches the fp16 success rate at \textbf{2.5$\times$} less memory, and on the bimanual platform it preserves long-horizon dual-arm skills that AWQ loses at the same budget. 
\end{abstract}

\begin{figure}[t]
\centering
\includegraphics[width=\textwidth]{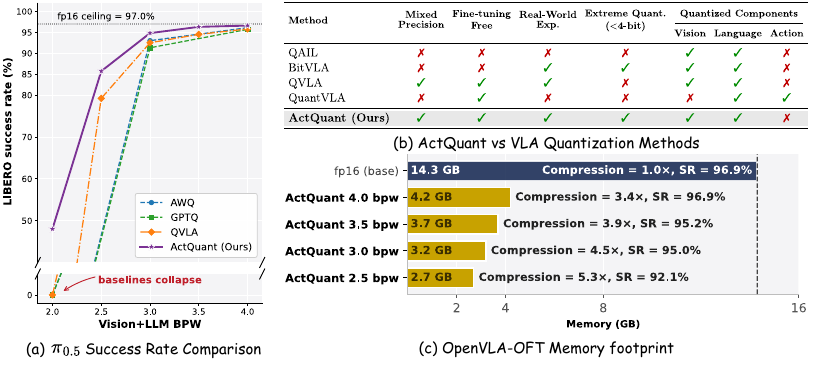}
\vspace{-17pt}
\caption{(a) Average success rate on LIBERO ($\pi_{0.5}$) as the backbone bit-width drops from $4.0$ to $2.0$ bpw. (b) Comparison of ActQuant with other VLA quantization methods. (c) Backbone memory usage across average bit-widths, with compression ratio against the baseline and success rate (SR).}
\label{fig:method-classification}
\end{figure}

\section{Introduction}
\label{sec:intro}

Vision-Language-Action (VLA) models~\citep{zhong2025survey} have become the dominant approach to generalist robotic manipulation, mapping visual observations and language directly into low-level motor commands. Their effectiveness is tightly coupled to the scale of \emph{both} the language backbone and the vision encoder. OpenVLA~\citep{openvla} and follow-ups~\citep{oft, li2024cogact}, pioneered discrete action generation by encoding action tokens directly into the LLaMA2 vocabulary\citep{touvron2023llama}. A second line, the $\pi$ family models~\citep{pi0,pi}, instead attaches a flow-matching action head for continuous action generation.  As VLA models scale toward general-purpose robotic foundation models, edge deployment on the robot itself becomes load-bearing: OpenVLA-OFT alone is built on a 7B LLM  plus a vision encoder, occupying $\sim$14 GB of weight memory in fp16, while on-board accelerators such as the NVIDIA Jetson Orin family expose only 8--16,GB of unified memory shared across perception and control. Aggressive quantization is therefore the natural lever for closing the gap between model size and memory.

The sub-4-bit regime exposes an inversion of a familiar pattern in machine learning systems: algorithmic advances, such as quantization and pruning, usually outpace system support, with deployment efficiency lagging behind model innovation~\citep{gptq,awq,ma2021non}.   
Intriguingly, mainstream on-device runtimes built on the GGML tensor library~\cite{ggml}(e.g., \textit{llama.cpp}~\citep{llamacpp}) already ship block-quantization kernels that make 2--4-bit inference practical for LLMs, potentially transferable to VLA models. However, algorithmic methods struggle to preserve model quality once pushed below the 4-bit threshold. The gap is particularly pronounced for VLA models, where even small quantization errors can cascade through actions and existing methods degrade sharply below 3\,bpw, losing the task entirely at 2.0\,bpw (Figure~\ref{fig:method-classification}(a)), leaving a significant portion of the efficiency potential offered by modern inference systems underutilized in practice.

Existing post-training quantization (PTQ) methods are mainly dedicated to LLMs~\citep{gptq, awq, squeezellm, tseng2024qtip} and do not transfer cleanly to VLA models. VLA performance depends on the quantized backbone preserving features that a downstream action head can decode into control signals. Meanwhile, layer-wise sensitivity-based quantization methods   ~\citep{gptq,awq} optimize a proxy only loosely connected to action prediction. LLM-derived quantization methods calibrate weight importance against the language-modeling signal alone. However, many VLA models are trained with a second supervisory pathway at the action head, and quantization quality depends on errors along both pathways rather than on the language signal in isolation~\citep{openvla, oft, li2024cogact}. Critically, quantization error does not vanish at a single output; it compounds over the closed-loop rollout, pushing the robot into states the calibration set never covered. These properties raise a question that current existing quantization methods do not resolve: \textit{how can we quantize VLA backbones below four bits while preserving task success rate?}

Mixed-precision quantization~\cite{ompq, spqr, chen2024channel, slim} seems to be a promising solution to extreme low-bit accuracy loss, but existing recipes sit at the wrong granularity for hardware-friendly VLA deployment. Both codebook-based schemes~\citep{aqlm, tseng2024qtip} and channel-wise quantization methods~\citep{qvla, spqr} yield irregular memory access patterns that are hardware-unfriendly for standard low-bit GPU kernels~\citep{yao2021hawq} (further discussed in Section~\ref{sec:related}). On the other hand, layer-wise allocation~\cite{hawqv2, mxprecision} keeps a single bit-width across each layer and stays hardware-friendly, but is too coarse to track the sensitivity variations that dominate below four bits. This leaves a per-tensor middle ground unexplored, where bit-widths vary across matrices to match action-relevance but stay uniform within each matrix so one dense integer kernel suffices for dequantization, all while preserving accuracy in the sub-4-bit regime. 

We propose ActQuant, an action-guided hardware-friendly quantization framework that allocates precision at two complementary granularities, both keyed to how much each weight contributes to predicting the VLA model's actions. Our method has two key steps: (1) \textbf{Inter-Tensor Bit Allocation}, which assigns each weight matrix its own bit-width based on how much it contributes to action prediction, giving more bits to action-critical matrices and fewer bits to redundant ones; and (2) \textbf{Intra-Tensor Scale Optimization}, where every block within a matrix keeps the bit-width assigned in step (1), but its scale is re-optimized using an action-aware importance signal. 
% This redistributes dynamic range toward the weights that most influence control without modifying the quantization bins themselves.
ActQuant keeps a single bit-width within each weight matrix, letting every matmul dispatch to one dense integer kernel.

While ActQuant is a hardware-friendly quantization, without efficient low-bit kernels at inference time it cannot translate into practical edge deployment. We therefore further introduce \textbf{OmniModel.cpp}, an agentic pipeline that converts diverse VLA models into a native C/C++ inference build with no PyTorch dependency, so ActQuant-quantized models run directly and efficiently on the robot. Our contributions are as follows:
\begin{itemize}
  \item We introduce ActQuant, an action-guided mixed-precision PTQ framework with two stages. Inter-Tensor Bit Allocation assigns each weight matrix its own bit-width based on how much that matrix contributes to action generation, and Intra-Tensor Scale Optimization refines the per-block scales within each matrix using an action-aware importance signal.
  \item We further introduce OmniModel.cpp, an agentic pipeline that delivers the on-device benefits of ActQuant by automatically converting a PyTorch VLA model into an equivalent native C/C++ inference implementation built on GGML's existing low-bit kernels. We integrate ActQuant directly into OmniModel.cpp, achieving up to $1.5\times$ lower per-token latency.
    \item Extensive experiments on LIBERO and on single-arm and bimanual UR3 platforms show that ActQuant compresses OpenVLA-OFT by 5.3$\times$ at 90.1\% closed-loop success at 2.5 bits-per-weight, where QVLA scores 0.0\%, and that on $\pi_{0.5}$ it is the only evaluated method still functional at 2.0 bits-per-weight (Figure~\ref{fig:method-classification}(b)). On the UR3 arm, we further deploy $\pi_{0.5}$ at 2.5$\times$ less memory while retaining the success rate.
   % , as summarized in Figure~\ref{fig:method-classification}(b).
\end{itemize}

\section{Related Work}
\label{sec:related}

\paragraph{Vision-Language-Action Models.}
Vision-Language-Action (VLA) models map perception and language instructions into low-level motor commands through a single foundation-scale backbone. RT-2~\cite{rt2} and OpenVLA~\cite{openvla} introduced the transformer-based VLA paradigm by discretizing the continuous action space and casting control as next-token prediction over a vision-language backbone. Building on this template, subsequent models such as OpenVLA-OFT~\cite{oft} and CogACT~\cite{li2024cogact} attach continuous action heads on top of the OpenVLA family to generate continuous control signals directly. A second line of work targets end-to-end continuous action generation by replacing the discrete action head with a flow-matching action expert; $\pi_{0}$~\cite{pi0} and $\pi_{0.5}$~\cite{pi} are the representative members. Although these models have produced a substantial leap in generalist action generation, they carry immense computational footprint~\cite{tinyvla,efficientvla}. Consequently, as VLA backbones continue to scale, extreme quantization has become a necessity for edge deployment on robotic platforms.

\paragraph{Post-Training Quantization.}
Quantization for VLA models is an emerging area with a small set of representative methods. BitVLA~\cite{bitvla} trains a native $1$-bit VLA from scratch with a ternary-weight backbone and a distillation-aware quantized vision encoder, while QAIL~\cite{qail} applies quantization-aware training to an imitation-learning policy at uniform $4$-bit weights and activations. Both rely on a full training pipeline and are therefore not training-free post-hoc methods. QuantVLA~\cite{quantvla} is a training-free framework that targets \emph{weight-activation} quantization (4-bit weights, 8-bit activations) for VLA models with a DiT-style action head, using attention-temperature matching and output-head balancing to recover scales after quantization. QVLA~\cite{qvla} allocates bit-widths at channel (per-row) granularity using a Taylor proxy of action-space sensitivity, which is too coarse to recover accuracy below four bits and is hardware-unfriendly~\cite{yao2021hawq}. None of these methods explores the extreme weight-only sub-4-bit regime that has been thoroughly studied for pure LLMs (e.g., SliM-LLM~\cite{slim}, QTIP~\cite{tseng2024qtip}, and AQLM~\cite{aqlm}). Our work fills exactly this gap with a hardware-friendly, mixed-precision aggressive weight quantization recipe for VLA models, as summarized in Figure~\ref{fig:method-classification}(b).

\section{Preliminaries}
\label{sec:background}

\paragraph{Vision-Language-Action models.}
VLA models typically decompose into four modules: a vision encoder $\mathcal{E}_v$, a projector $\mathcal{P}$ into the language embedding space, a transformer-based language backbone $\mathcal{G}$, and an action head $\mathcal{H}$ that produces the deployed control signal. Since $\mathcal{E}_v$ and $\mathcal{G}$ hold the majority of the parameter mass while $\mathcal{P}$ and $\mathcal{H}$ are compact yet highly sensitive to perturbation~\cite{qvla}, we quantize only the weights of $\mathcal{E}_v$ and $\mathcal{G}$ and keep $\mathcal{P}$ and $\mathcal{H}$ in full precision. Both $\mathcal{E}_v$ and $\mathcal{G}$ are generally stacks of $L$ transformer blocks indexed by $\ell = 1, \ldots, L$, each containing attention projections $\{W_Q^{(\ell)}, W_K^{(\ell)}, W_V^{(\ell)}, W_O^{(\ell)}\}$ and an MLP block with matrices $\{W_{\mathrm{up}}^{(\ell)}, W_{\mathrm{down}}^{(\ell)}, W_{\mathrm{gate}}^{(\ell)}\}$. For brevity, we collect the module indices in $\mathcal{J} = \{Q, K, V, O, \mathrm{up}, \mathrm{down}, \mathrm{gate}\}$, denote any of these matrices as $W_j^{(\ell)}$ with $j \in \mathcal{J}$, and treat each $W_j^{(\ell)}$ as an independent quantization candidate.

\paragraph{Weight quantization.}
We adopt the block quantization format as exemplified by QLoRA~\cite{dettmers2023qlora}, in which the elements of each weight matrix $W_j^{(\ell)}$ are organized into a \emph{block} / \emph{super-block} hierarchy. Every $B$ consecutive elements form a block $b$ with element index set $\mathcal{B}_b = \{1, \ldots, B\}$, its own scale $s_b \in \mathbb{R}$, and its own zero point $z_b \in \mathbb{R}$ (the symmetric special case sets $z_b = 0$); every $S$ consecutive blocks then form a super-block $\mathcal{S}$ with its own scale $S_{\mathcal{S}}$ and zero point $Z_{\mathcal{S}}$ that compress the per-block $(s_b, z_b)$ themselves. For an $n$-bit quantizer, every element in block $b$ is encoded with the same $n$ bits and dequantized as
\begin{equation}
\label{eq:weight-quant}
\hat{w}_i \;=\; s_b\,(q_i - z_b), \quad q_i \;=\; \operatorname*{\arg\min}_{q \in \mathcal{Q}_n}\, \bigl|\, w_i - s_b\,(q - z_b)\,\bigr|, \quad i \in \mathcal{B}_b,
\end{equation}
where $\mathcal{Q}_n$ is a format-specific codebook of representable values which can be uniform or non-uniform. The per-block parameters $(s_b, z_b)$ and the assignments $\{q_i\}_{i \in \mathcal{B}_b} \subset \mathcal{Q}_n$ are jointly chosen to minimize a per-block reconstruction error. We denote the quantized counterpart of any weight matrix $W_j^{(\ell)}$ as $\hat{W}_j^{(\ell)}$, with elements $\hat{w}_i$ defined in Eq.~\eqref{eq:weight-quant}.

\paragraph{Hilbert-Schmidt Independence Criterion (HSIC).}
\label{sec:hsic-bg}
The Hilbert-Schmidt Independence Criterion (HSIC)~\cite{gretton2005measuring} is a kernel-based measure of statistical dependence between two random variables that, unlike mutual information, is tractable to compute and requires no density estimation. Given $K$ i.i.d.\ samples $\{(a_k, b_k)\}_{k=1}^{K}$ drawn from $(A, B)$ and kernel functions $k_A, k_B$, HSIC is estimated empirically as
\begin{equation} \label{eq:hsic-emp}
\widehat{\mathrm{HSIC}}(A, B) \;=\; (K-1)^{-2}\, \operatorname{tr}\!\left(\mathbf{K}_A\, \mathbf{C}\, \mathbf{K}_B\, \mathbf{C}\right),
\end{equation}
where $\mathbf{K}_A, \mathbf{K}_B \in \mathbb{R}^{K \times K}$ are kernel matrices with entries $[\mathbf{K}_A]_{ij} = k_A(a_i, a_j)$ and $[\mathbf{K}_B]_{ij} = k_B(b_i, b_j)$, and $\mathbf{C} = \mathbf{I} - \tfrac{1}{K}\mathbf{1}\mathbf{1}^{\top}$ is the centering matrix. We use the RBF kernel throughout, under which $\widehat{\mathrm{HSIC}}(A,B) \to 0$ if and only if $A \perp B$, making HSIC a faithful dependence measure. The population form, the universal-kernel guarantee, and prior uses of HSIC as an information-bottleneck surrogate are reviewed in Appendix~\ref{app:hsic}.

\paragraph{Hessian-based sensitivity and Fisher approximation.}
Let $\boldsymbol{\theta} \in \mathbb{R}^P$ collect all trainable parameters of the quantization-target modules $\mathcal{E}_v$ and $\mathcal{G}$, with $P = \sum_{\ell=1}^{L} \sum_{j \in \mathcal{J}} |W_j^{(\ell)}|$, and let $\mathcal{L}(\boldsymbol{\theta})$ be the action loss averaged over a calibration distribution drawn from the deployed task. Fine-tuned VLA backbones operate near a local minimum of $\mathcal{L}$, hence $\nabla_{\boldsymbol{\theta}}\mathcal{L} \approx \mathbf{0}$, and a second-order Taylor expansion reduces the deployment objective to a Hessian-weighted reconstruction problem (full derivation in Appendix~\ref{app:hessian}),
\begin{equation} \label{eq:quant-obj}
\boldsymbol{\hat{\theta}}^{\star} \;=\; \operatorname*{\arg\min}_{\boldsymbol{\hat{\theta}}}\; (\hat{\boldsymbol{\theta}} - \boldsymbol{\theta})^{\top}\, \mathbf{H}\, (\hat{\boldsymbol{\theta}} - \boldsymbol{\theta}),
\end{equation}
where $\mathbf{H} = \nabla^{2}_{\boldsymbol{\theta}}\mathcal{L}$ is the Hessian of the action loss and the diagonal entry $H_{ii}$ measures the curvature of $\mathcal{L}$ along $\theta_i$.Computing $\mathbf{H}$ exactly requires second derivatives at backbone scale, which is prohibitively expensive, so we follow standard practice~\cite{martens2020new, squeezellm} and approximate it by the empirical Fisher information matrix
\begin{equation} \label{eq:fisher-bg}
\mathcal{F}(\boldsymbol{\theta}) \;=\; \frac{1}{|\mathcal{D}|}\sum_{d \in \mathcal{D}} \nabla_{\boldsymbol{\theta}} \mathcal{L}_d \,\bigl(\nabla_{\boldsymbol{\theta}} \mathcal{L}_d\bigr)^{\top},
\end{equation}
which agrees with the Hessian at a local minimum under mild regularity conditions (Appendix~\ref{app:hessian}). Storing the full $\mathcal{F} \in \mathbb{R}^{P \times P}$ is infeasible at this parameter scale, so in Section~\ref{sec:scale-opt} we treat each weight matrix $W_j^{(\ell)}$ as an independent quantization target and use the diagonal approximation of $\mathcal{F}$ as a per-element importance score.

\section{Methodology}
\label{sec:method}

\begin{figure}[t]
\centering
\includegraphics[width=\textwidth]{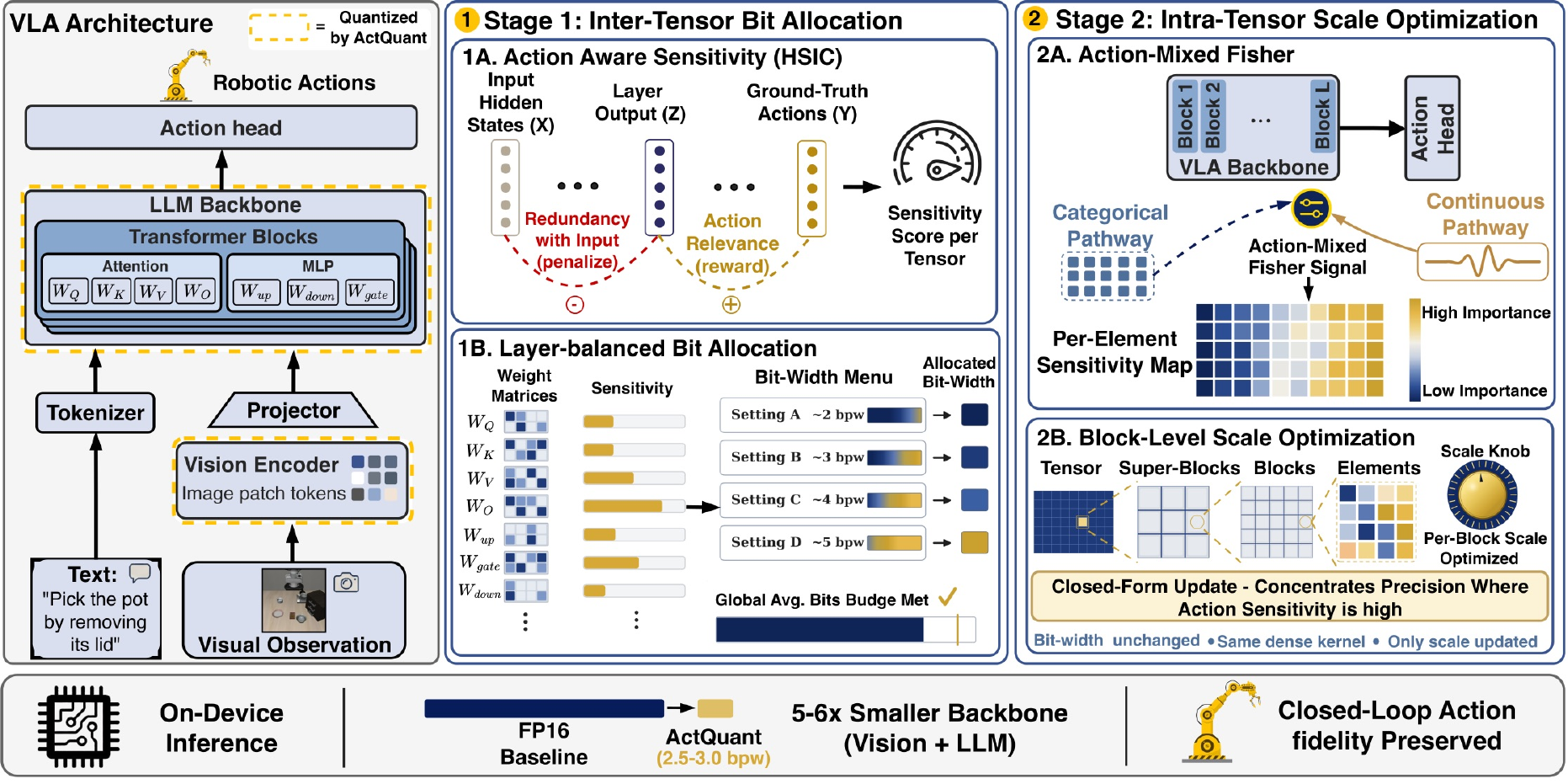}
\caption{\textbf{Overview of ActQuant.} \textbf{Stage 1 (inter-tensor):} an action-aware HSIC sensitivity score assigns each matrix a single bit-width under a budget. \textbf{Stage 2 (intra-tensor):} with the bit-width fixed, the per-block scales within each matrix are optimized, weighted by per-element sensitivities from an Action-Mixed Fisher that combines the action-head loss with an optional LM loss.}
\label{fig:main}
\end{figure}

Building on the formulation introduced in Section~\ref{sec:background}, we propose ActQuant, which decomposes weight quantization on the VLA backbone into two action-guided allocation problems, illustrated end-to-end in Figure~\ref{fig:main}. Both LLM-derived and VLA-specific quantization methods significantly hurt closed-loop action quality on modern VLA backbones in the sub-4-bit regime (Figure~\ref{fig:method-classification}a). Therefore, recovering accuracy at this scale requires an \emph{action-guided}, mixed-precision compression method that remains hardware-friendly. 
% Fine-grained schemes that vary the bit-width per element, per row, or per channel induce irregular memory access patterns which is not supported efficiently . 
We propose two complementary, hardware-friendly procedures that together answer two questions: \textbf{(Q1)}~how can we minimize quantization error on the tensors that contribute most to action generation while keeping precision uniform within each tensor? \textbf{(Q2)}~while preserving that tensor-level uniformity, how do we stay fine-grained enough to retain task success rate in the sub-4-bit regime?

Section~\ref{sec:hsic} answers \textbf{Q1} with a novel HSIC-based sensitivity score that allocates one bit-width per tensor according to how much it contributes to predicting the ground-truth action. Section~\ref{sec:scale-opt} answers \textbf{Q2} with a novel VLA-specific per-element sensitivity metric, the Action-Mixed Fisher (AMF), which is being used to further optimize the per-block scales and zero points rather than changing the bit assignment. Section~\ref{sec:omnimodel} then describes how we deploy the resulting quantized models on edge hardware through OmniModel.cpp.
% This way, fine-grained dynamic-range adjustment is achieved without disturbing the hardware-friendly tensor-level layout. Both signals are computed once on a shared observation-action calibration set.

%-------------------------------------------------------------------
\subsection{Inter-Tensor Bit Allocation}
\label{sec:hsic}
%-------------------------------------------------------------------

We need a per-tensor sensitivity score that is large for matrices whose output carries action-relevant information and small for matrices that can be quantized aggressively without hurting the action generation quality.

\paragraph{Action-aware sensitivity.}
Let $X \in \mathbb{R}^{K \times d_{\mathrm{in}}}$ collect the input hidden states of the backbone over the calibration set $\mathcal{D}$ with $K = |\mathcal{D}|$ samples, $Z_j^{(\ell)} \in \mathbb{R}^{K \times d_j^{(\ell)}}$ the corresponding output hidden states of $W_j^{(\ell)}$, and $Y \in \mathbb{R}^{K \times d_a}$ the ground-truth action labels. We define the sensitivity of $W_j^{(\ell)}$ as
\begin{equation}
\label{eq:hsic-sens}
\mathrm{S}\!\left(W_j^{(\ell)}\right) \;=\; -\,\lambda_{\mathrm{red}}\,\widehat{\mathrm{HSIC}}\!\left(X,\, Z_j^{(\ell)}\right) \;+\; \lambda_{\mathrm{act}}\,\widehat{\mathrm{HSIC}}\!\left(Z_j^{(\ell)},\, Y\right),
\end{equation}
where $\widehat{\mathrm{HSIC}}$ is the empirical estimator from Eq.~\eqref{eq:hsic-emp}, and $\lambda_{\mathrm{red}}, \lambda_{\mathrm{act}} \in \mathbb{R}_{+}$ are balancing hyperparameters. The second term, weighted by $\lambda_{\mathrm{act}}$, is a \emph{task-relevance} score: it grows when the layer's output is statistically aligned with the ground-truth actions $Y$, marking the layer as critical for closed-loop control and therefore deserving of more bits. The first term, weighted by $-\lambda_{\mathrm{red}}$, measures \emph{input redundancy}: a large value indicates the layer's output carries redundant or noisy information contained in $X$, so this layer can absorb more quantization error. We subtract this term so that two matrices with identical task relevance are differentiated by how much new information they actually inject into the residual stream. The coefficients $\lambda_{\mathrm{red}}$ and $\lambda_{\mathrm{act}}$ rebalance the two HSIC scales, which can differ by orders of magnitude depending on the calibration set and kernel bandwidth, and let us emphasize task relevance over input redundancy (or vice versa) without re-scaling the kernels themselves.

\paragraph{Layer-balanced bit allocation.}
Turning the sensitivities $\mathrm{S}(W_j^{(\ell)})$ into a bit-allocation requires pairing each tensor with the cost of quantizing it at a given precision. Following the classical quantization theory~\cite{lloyd1982least}, the per-element squared error of each type $t \in \mathcal{T}$ scales as $\eta(t) = 2^{-2\,b(t)}$. Therefore, the predicted reconstruction error of $W_j^{(\ell)}$ at type $t_j^{(\ell)}$ is $\mathrm{S}(W_j^{(\ell)})\,\eta(t_j^{(\ell)})$. Aggregating these errors linearly across the whole backbone would let the optimizer absorb a large error in a single sensitive tensor in exchange for many small savings elsewhere, exactly the failure mode that collapses the task success rate. We instead group the per-tensor error by transformer layer and minimize the sum of squared per-layer errors under an average bits-per-weight budget,
\begin{equation}
\label{eq:hsic-alloc}
\begin{aligned}
&\underset{\{t_j^{(\ell)}\}_{\ell,\,j}}{\min}\; \sum_{\ell=1}^{L}\, E_\ell^{2}, \qquad E_\ell \;=\; \sum_{j \in \mathcal{J}}\, \mathrm{S}(W_j^{(\ell)})\, \eta\!\left(t_j^{(\ell)}\right),
&\text{s.t.}\quad \frac{\sum_{\ell, j}\, b\!\left(t_j^{(\ell)}\right)\, |W_j^{(\ell)}|}{\sum_{\ell, j}\, |W_j^{(\ell)}|} \;\le\; \bar{b}.
\end{aligned}
\end{equation}
The squared sum penalizes large per-layer errors disproportionately, forcing every linear layer to stay within a comparable error band rather than sacrificing one layer for two others. We revisit this choice in the ablation study in Section~\ref{sec:exp-ablation}.

\paragraph{Solver.}
Eq.~\eqref{eq:hsic-alloc} is a discrete bit-allocation problem with one categorical choice per tensor and one global budget; standard greedy and Lagrangian-relaxation methods solve it with bounded suboptimality~\cite{everett1963generalized,nemhauser1978analysis}. We adopt a greedy procedure that starts every tensor at the cheapest type and, at each step, applies the upgrade that maximizes the gain-per-bit ratio $(E_\ell^{2} - \tilde{E}_\ell^{2})/\Delta\mathrm{bits}$, where $\tilde{E}_\ell$ is the layer error after the candidate upgrade and $\Delta\mathrm{bits}$ the extra bits it costs. The loop stops once the budget is met.

%-------------------------------------------------------------------
\subsection{Intra-Tensor Scale Optimization}
\label{sec:scale-opt}
%-------------------------------------------------------------------

With the per-tensor types $\{t_j^{(\ell)}\}$ from Section~\ref{sec:hsic} fixed, we choose the per-block scales $\{s_b\}$ \emph{within} each $W_j^{(\ell)}$. As shown in Eq.~\eqref{eq:quant-obj}, this objective is Hessian-weighted; we employ the empirical Fisher information matrix of Eq.~\eqref{eq:fisher-bg} as a tractable approximation to the Hessian, and, assuming that cross-weight interactions are negligible, use its diagonal elements as a per-element importance score. ActQuant tailors this Fisher signal to reflect the deployed action loss.

\paragraph{Action-Mixed Fisher (AMF).}
Many architectures, such as OpenVLA-OFT, additionally expose a second pathway through a language-modeling head $W_{\mathrm{lm}}$ that outputs logits over a discrete action vocabulary $\mathcal{V}$ at the sequence positions reserved for action tokens. Let $\mathcal{I}_d^{\mathrm{cls}}$ denote those action-token positions in calibration sample $d$ and $y_{d, p} \in \mathcal{V}$ the ground-truth action token at position $p$, giving the cross-entropy loss,
\begin{equation}
\label{eq:cls-loss}
\mathcal{L}^{\mathrm{cls}}_d(\boldsymbol{\theta}) \;=\; -\frac{1}{|\mathcal{I}_d^{\mathrm{cls}}|} \sum_{p \in \mathcal{I}_d^{\mathrm{cls}}} \log \mathrm{softmax}\!\bigl(W_{\mathrm{lm}}\, \mathbf{h}_{d, p}(\boldsymbol{\theta})\bigr)_{y_{d, p}},
\end{equation}
where $\mathbf{h}_{d, p}(\boldsymbol{\theta})$ is the backbone hidden state at position $p$ for sample $d$. We define the \emph{Action-Mixed Fisher} loss as the convex combination
\begin{equation}
\label{eq:amf-loss}
\mathcal{L}^{\mathrm{AMF}}_d(\boldsymbol{\theta}) \;=\; \alpha\, \mathcal{L}^{\mathrm{act}}_d(\boldsymbol{\theta}) \;+\; (1 - \alpha)\, \mathcal{L}^{\mathrm{cls}}_d(\boldsymbol{\theta}), \qquad \alpha \in [0, 1].
\end{equation}
This formulation makes AMF apply uniformly to both kinds of VLA models: for architectures with both pathways, such as OpenVLA-OFT, $\alpha \in (0, 1)$ blends the action-head and categorical signals, while for architectures with only the action-head pathway, such as $\pi_{0.5}$, we set $\alpha = 1$ so the second term drops out and AMF reduces to the Action-Only Fisher. Applying the diagonal approximation to the corresponding AMF Fisher matrix yields the per-element scores
\begin{equation}
\label{eq:amf-fisher}
\mathcal{F}^{\mathrm{AMF}}_{ii} \;=\; \frac{1}{|\mathcal{D}|}\, \sum_{d \in \mathcal{D}} \left(\frac{\partial \mathcal{L}^{\mathrm{AMF}}_d}{\partial \theta_i}\right)^{2}, \qquad i = 1, \ldots, P,
\end{equation}
computed from one backward pass per calibration sample. The action-head signal anchors AMF to the deployed action quality, and mixing in the categorical signal, when available, additionally reflects the joint distribution over discrete action categories that the LM head exposes. Combining the two retains both views and additionally exposes a cross-pathway covariance term (Appendix~\ref{app:cross-cov}) that single-loss Fishers do not capture.

\paragraph{Block-Level Scale Optimization.}
Substituting $\mathbf{H} \approx \mathrm{diag}(\mathcal{F}^{\mathrm{AMF}})$ in Eq.~\eqref{eq:quant-obj} decouples the joint quadratic into independent per-block problems. Our per-element weight sensitivity $\omega_{b,i}$ multiplies the Action-Mixed-Fisher sensitivities by a local weight magnitude factor, so that quantization noise is penalized most heavily on elements that are simultaneously task-critical and large within their block. For block $b$ with element index set $\mathcal{B}_b$ and original weights $\{w_i\}_{i \in \mathcal{B}_b}$,
\begin{equation}
\label{eq:scale-opt}
\omega_{b,i} \;=\; \mathcal{F}^{\mathrm{AMF}}_{ii}\,\sqrt{\sigma_b^{2} + w_i^{2}},
\qquad
(s_b^{\star},\, \mathbf{q}_b^{\star}) \;=\; \operatorname*{\arg\min}_{s_b \in \mathbb{R},\; q_i \in \mathcal{Q}_n}\; \sum_{i \in \mathcal{B}_b} \omega_{b,i}\, (w_i - s_b\, q_i)^{2},
\end{equation}
where $\sigma_b^{2} = |\mathcal{B}_b|^{-1}\!\sum_{j \in \mathcal{B}_b} w_j^{2}$ is the mean squared weight in the block; we denote the inner sum by $\Phi_b(s_b, \mathbf{q}_b)$. The objective $\Phi_b$ is non-convex in $(s_b, \mathbf{q}_b)$ but becomes tractable when one variable is fixed: with $s_b$ fixed, each codeword $q_i$ is chosen independently as the codebook entry closest to the rescaled weight $w_i/s_b$, and the importance weight $\omega_{b,i}$ drops out because the minimization decouples across $i$; with $\{q_i\}$ fixed, $\Phi_b$ becomes a weighted least-squares problem in $s_b$ alone whose stationarity condition $\partial \Phi_b / \partial s_b = 0$ admits a closed form. We therefore alternate the two steps,
\begin{equation}
\label{eq:scale-update}
q_i^{\star} \;=\; \operatorname*{\arg\min}_{q \in \mathcal{Q}_n}\, \bigl|\, w_i / s_b - q\,\bigr|, \qquad s_b^{\star} \;=\; \frac{\sum_{i \in \mathcal{B}_b} \omega_{b,i}\, w_i\, q_i^{\star}}{\sum_{i \in \mathcal{B}_b} \omega_{b,i}\, \left(q_i^{\star}\right)^{2}},
\end{equation}
which is the constrained analogue of Lloyd-Max scalar quantization specialized to a fixed codebook $\mathcal{Q}_n$ and a learnable scale; the iterates monotonically decrease $\Phi_b$ and converge to a local minimum in a few rounds. 

%-------------------------------------------------------------------
\subsection{Deployment with OmniModel.cpp}
\label{sec:omnimodel}
%-------------------------------------------------------------------

Aggressive quantization only translates into edge-deployment gains when the quantized model can be executed by efficient low-bit kernels. The GGML tensor library~\cite{ggml} together with \texttt{llama.cpp}~\cite{llamacpp} already ships hand-written low-bit kernels that make sub-4-bit inference practical, but their support is limited to a narrow set of decoder-only LLMs. To bridge this gap, we introduce \textbf{OmniModel.cpp}, an agentic pipeline that automatically rewrites a PyTorch VLA model as a native C/C++ inference program expressed in the GGML computation graph. We pair OmniModel.cpp with ActQuant so that quantized VLA architectures can be deployed and run efficiently on edge hardware. Because the converted model now lives inside this graph, its matrix multiplies dispatch directly to GGML's existing low-bit kernels, and ActQuant's per-tensor bit-widths and per-block scales translate immediately into reduced memory footprint and latency. We describe the pipeline's planner, agent layers, and verification loop in detail in Appendix~\ref{app:omnimodel}.

\section{Experiments}
\label{sec:experiments}
\subsection{Experimental Setup}
\label{sec:exp-setup}

\begin{figure}[t]
    \centering
    \includegraphics[width=\linewidth]{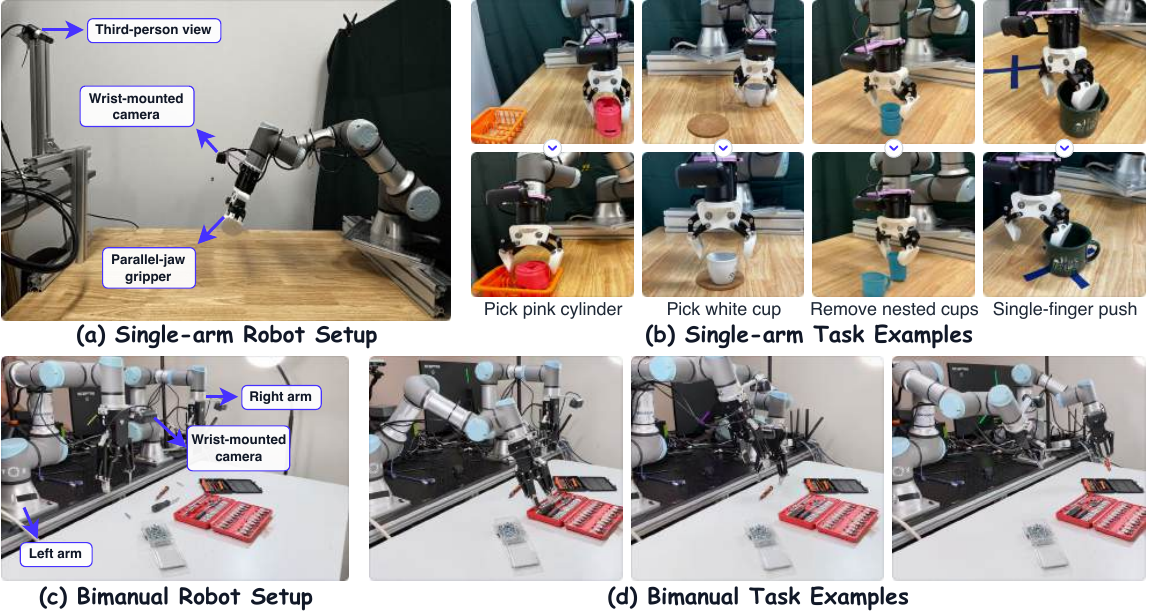}
    \vspace{-21pt}
    \caption{%
        Physical robot setup and real-world task demonstrations.
        The system consists of a UR3 robot arm and a bimanual UR3 setup.
        % \textbf{(b)} Real-world manipulation tasks including pick-and-place, 
        % cup unstacking, and object pushing.
    }
    \label{fig:robot_setup}
\end{figure}

We use the widely adopted LIBERO benchmark~\cite{liu2023libero}, which provides four object-manipulation suites, and two real-world Universal Robots UR3 platforms: a single $6$-DoF arm equipped with a wrist-mounted RGB camera and a third-person Intel RealSense D435i camera (Figure~\ref{fig:robot_setup}), and a bimanual UR3 platform used for long-horizon dual-arm manipulation. We evaluate two state-of-the-art VLA policies with distinct action-prediction architectures, OpenVLA-OFT~\cite{oft} and $\pi_{0.5}$\cite{pi}, and compare ActQuant against three weight-only PTQ baselines, AWQ\cite{awq}, GPTQ~\cite{gptq}, and the only VLA mixed-precision weight quantization method, QVLA~\cite{qvla}. All methods share the same calibration set of only $60$ episodes drawn from the LIBERO fine-tuning split (Appendix~\ref{app:calibration}). We sweep average backbone bits-per-weight in $\{4.0,\, 3.5,\, 3.0,\, 2.5\}$ for OpenVLA-OFT and additionally $2.0$,bpw for $\pi_{0.5}$, with the vision encoder and the language backbone jointly held at the same target bit-width. Models quantized with ActQuant are fully integrated with OmniModel.cpp (Appendix~\ref{app:omnimodel}); hardware and runtime details are in Appendix~\ref{app:hardware}.

\subsection{Experimental Results}
\label{sec:exp-results}

\begin{table}[t]
\centering
\caption{Comparison of quantization methods on VLA models. RTN, AWQ, and GPTQ use uniform integer precision, while QVLA and \textbf{Ours} support mixed-precision (avg. bpw). }
\label{tab:vla-quant-results}
\resizebox{0.84\textwidth}{!}{%
\begin{tabular}{l| l| c |ccccc| c| c c}
\toprule
\multirow{2}{*}{\textbf{Model}} & \multirow{2}{*}{\textbf{Method}} & \multirow{2}{*}{\shortstack{\textbf{Vision+LLM}\\\textbf{BPW}}} & \multicolumn{5}{c}{\textbf{Success Rate (\%)}} & & \multirow{2}{*}{\shortstack{\textbf{VLA}\\\textbf{BPW}}} & \multirow{2}{*}{\shortstack{\textbf{Mem.}\\\textbf{(GB)}}}  \\
\cmidrule(lr){4-8}
 &  &  & \textbf{Spatial} & \textbf{Object} & \textbf{Goal} & \textbf{Long} & \textbf{Avg.} & \textbf{$\Delta$} &  &  \\
\midrule
\multirow{20}{*}{\shortstack{\textbf{OpenVLA-}\\\textbf{OFT}}} & Baseline & 16.0 & 97.6 & 98.4 & 96.8 & 95.1 & 96.9 & 0.00 & 16.0 & 14.3 \\
% \cmidrule(lr){2-11}
% & \multirow{3}{*}{RTN} & 4.0 & -- & -- & -- & -- & -- & -- & 4.x & -- \\
% &                      & 3.0 & -- & -- & -- & -- & -- & -- & 3.x & -- \\
\cmidrule(lr){2-11}
& \multirow{3}{*}{AWQ} & 4.0 &  \underline{94.6} & 98.6 & 96.8 & 94.4 & 96.1 & $-0.8$ & 4.3 & 4.1 \\
&                      & 3.0 & \underline{86.4} & \textbf{98.0} & \underline{90.2} & \underline{91.4} & \underline{91.5} & $-5.4$ & 3.4 & 3.2 \\
&                      & 2.0 & 0.0 & 0.0 & 0.0 & 0.0 & 0.0 & $-96.9$ & 2.7 & 2.4 \\
\cmidrule(lr){2-11}
& \multirow{3}{*}{GPTQ} & 4.0 & 90.8 & \textbf{98.8} & 96.4 & 92.6 & 94.6 & $-2.3$ & 4.3 & 4.1 \\
&                       & 3.0 & 82.4 & \textbf{98.0} & 88.8 & 87.2 & 89.0 & $-7.9$ & 3.4 & 3.2 \\
&                       & 2.0 & 0.0 & 0.0 & 0.0 & 0.0 & 0.0 & $-96.9$ & 2.7 & 2.4 \\
\cmidrule(lr){2-11}
& \multirow{4}{*}{QVLA} & 4.0 & \textbf{96.0} & \underline{98.7} & \textbf{97.4} & \textbf{95.2} & \textbf{96.8} & $-0.1$ & 4.8 & 4.5 \\
&                       & 3.5 & \underline{82.9} & \underline{97.2} & \underline{91.4} & \underline{93.6} & \underline{91.3} & $-5.6$ & 4.3 & 4.1 \\
&                       & 3.0 & 40.8 & 56.8 & 20.2 & 31.8 & 37.4 & $-59.5$ & 3.9 & 3.7 \\
&                       & 2.5 & 0.0 & 0.2 & 0.0 & 0.0 & 0.0 & $-96.9$ & 3.4 & 3.1 \\
\cmidrule(lr){2-11}
& \multirow{4}{*}{\textbf{ActQuant}} & \cellcolor{gray!15} 4.0 & \cellcolor{gray!15} \textbf{96.0} & \cellcolor{gray!15} 98.2 & \cellcolor{gray!15} \underline{97.2} & \cellcolor{gray!15} \underline{95.0} & \cellcolor{gray!15} \underline{96.6} & \cellcolor{gray!15} $-0.3$ & \cellcolor{gray!15} 4.4 & \cellcolor{gray!15} 4.2 \\
&                              & \cellcolor{gray!15} 3.5 & \cellcolor{gray!15} \textbf{95.8} & \cellcolor{gray!15} \textbf{98.4} & \cellcolor{gray!15} \textbf{96.0} & \cellcolor{gray!15} \textbf{95.6} & \cellcolor{gray!15} \textbf{96.5} & \cellcolor{gray!15} $-0.4$ & \cellcolor{gray!15} 3.8 & \cellcolor{gray!15} 3.4 \\
&                              & \cellcolor{gray!15} 3.0 & \cellcolor{gray!15} \textbf{92.8} & \cellcolor{gray!15} \underline{97.4} & \cellcolor{gray!15} \textbf{94.0} & \cellcolor{gray!15} \textbf{95.6} & \cellcolor{gray!15} \textbf{95.0} & \cellcolor{gray!15} $-1.9$ & \cellcolor{gray!15} 3.5 & \cellcolor{gray!15} 3.2 \\
&    & \cellcolor{gray!15} 2.5 & \cellcolor{gray!15} \textbf{86.4} & \cellcolor{gray!15} \textbf{98.2} & \cellcolor{gray!15} \textbf{84.8} & \cellcolor{gray!15} \textbf{91.0} & \cellcolor{gray!15} \textbf{90.1} & \cellcolor{gray!15} $-6.8$ & \cellcolor{gray!15} 3.0 & \cellcolor{gray!15} 2.7 \\
\midrule
\multirow{20}{*}{\scalebox{1.3}{\textbf{\boldmath$\pi_{0.5}$}}} & Baseline & 16.0 & 98.4 & 98.4 & 97.6 & 93.6 & 97.0 & 0.00 & 16.0 & 6.7 \\
\cmidrule(lr){2-11}
& \multirow{3}{*}{AWQ} & 4.0 & 96.6 & \underline{98.0} & \underline{96.8} & \textbf{93.0} & \underline{96.1} & $-0.9$ & 5.6 & 2.4 \\
&                      & 3.0 & 93.4 & 97.0 & \underline{91.4} & \textbf{90.2} & \underline{93.0} & $-4.0$ & 5.0 & 2.1 \\
&                      & 2.0 & 0.0 & 0.0 & 0.0 & 0.0 & 0.0 & $-97.0$ & 4.1 & 1.7 \\
\cmidrule(lr){2-11}
& \multirow{3}{*}{GPTQ} & 4.0 & 97.2 & 96.8 & \textbf{97.0} & \underline{91.8} & 95.7 & $-1.3$ & 5.6 & 2.4 \\
&                       & 3.0 & 92.2 & 95.4 & 88.6 & \underline{89.0} & 91.3 & $-5.7$ & 5.0 & 2.1 \\
&                       & 2.0 & 0.0 & 0.0 & 0.0 & 0.0 & 0.0 & $-97.0$ & 4.1 & 1.7 \\
% \cmidrule(lr){2-11}
\cmidrule(lr){2-11}
& \multirow{5}{*}{QVLA} & 4.0 & \underline{98.0} & 97.2 & 96.4 & \underline{91.8} & 95.8 & $-1.2$ & 6.3 & 2.7 \\
&                       & 3.5 & \underline{97.8} & \underline{96.2} & \underline{91.4} & \textbf{92.6} & \underline{94.5} & $-2.5$ & 6.0 & 2.5 \\
&                       & 3.0 & \underline{97.4} & \underline{97.8} & 87.8 & 86.8 & 92.5 & $-4.5$ & 5.6 & 2.4 \\
&                       & 2.5 & \underline{93.0} & \underline{97.6} & \underline{71.0} & \underline{55.0} & \underline{79.2} & $-17.8$ & 5.3 & 2.2 \\
&                       & 2.0 & 0.0 & 0.0 & 0.0 & 0.0 & 0.0 & $-97.0$ & 4.7 & 2.0 \\
\cmidrule(lr){2-11}
& \multirow{5}{*}{\textbf{ActQuant}} & \cellcolor{gray!15} 4.0 & \cellcolor{gray!15} \textbf{98.4} & \cellcolor{gray!15} \textbf{99.4} & \cellcolor{gray!15} \underline{96.8} & \cellcolor{gray!15} \underline{91.8} & \cellcolor{gray!15} \textbf{96.6} & \cellcolor{gray!15} $-0.4$ & \cellcolor{gray!15} 6.2 & \cellcolor{gray!15} 2.7 \\
&                              & \cellcolor{gray!15} 3.5 & \cellcolor{gray!15} \textbf{99.0} & \cellcolor{gray!15} \textbf{99.6} & \cellcolor{gray!15} \textbf{94.0} & \cellcolor{gray!15} \underline{92.4} & \cellcolor{gray!15} \textbf{96.3} & \cellcolor{gray!15} $-0.7$ & \cellcolor{gray!15} 5.8 & \cellcolor{gray!15} 2.5 \\
&                              & \cellcolor{gray!15} 3.0 & \cellcolor{gray!15} \textbf{98.2} & \cellcolor{gray!15} \textbf{98.8} & \cellcolor{gray!15} \textbf{95.0} & \cellcolor{gray!15} 87.2 & \cellcolor{gray!15} \textbf{94.8} & \cellcolor{gray!15} $-2.2$ & \cellcolor{gray!15} 5.6 & \cellcolor{gray!15} 2.4 \\
&                              & \cellcolor{gray!15} 2.5 & \cellcolor{gray!15} \textbf{96.6} & \cellcolor{gray!15} \textbf{98.4} & \cellcolor{gray!15} \textbf{74.0} & \cellcolor{gray!15} \textbf{73.6} & \cellcolor{gray!15} \textbf{85.7} & \cellcolor{gray!15} $-11.3$ & \cellcolor{gray!15} 5.2 & \cellcolor{gray!15} 2.2 \\
&                              & \cellcolor{gray!15} 2.0 & \cellcolor{gray!15} \textbf{61.2} & \cellcolor{gray!15} \textbf{80.0} & \cellcolor{gray!15} \textbf{25.0} & \cellcolor{gray!15} \textbf{25.0} & \cellcolor{gray!15} \textbf{48.0} & \cellcolor{gray!15} $-49.0$ & \cellcolor{gray!15} 4.8 & \cellcolor{gray!15} 2.0 \\
\bottomrule
\end{tabular}%
}
\end{table}

\begin{wraptable}{r}{0.52\textwidth}
\vspace{-20pt}
\centering
\caption{OpenVLA per-token latency (ms): PyTorch vs three OmniModel.cpp builds across three platforms. PyTorch lacks a CPU/GPU split (N/A).}
\label{tab:openvla-speed}
\resizebox{0.52\textwidth}{!}{%
\begin{tabular}{@{}llccc@{}}
\toprule
\multirow{2}{*}{\textbf{Impl.}} & \multirow{2}{*}{\textbf{Mode}} & \multicolumn{3}{c}{\textbf{Latency (ms)}} \\
\cmidrule(lr){3-5}
 & & \textbf{A6000} & \textbf{M4 Pro} & \textbf{AGX Thor} \\
\midrule
\multirow{2}{*}{PyTorch}
 & GPU & $233 \pm 1$    & $1319 \pm 49$  & $598.5 \pm 2$ \\
 & 50/50    & N/A            & N/A            & N/A \\
\midrule
\multirow{2}{*}{ours F16}
 & GPU & $260 \pm 2$    & $1269 \pm 57$  & $668 \pm 4$ \\
 & 50/50    & $1447 \pm 7$   & $3511 \pm 1822$ & $2602 \pm 29$ \\
\midrule
\multirow{2}{*}{ours Q8\_0}
 & GPU & $189$          & $1052 \pm 4$   & $529 \pm 7$ \\
 & 50/50    & $961 \pm 1$    & $1840 \pm 417$ & $2455 \pm 37$ \\
\midrule
\multirow{2}{*}{Ours Q4\_K\_M}
 & GPU & $\mathbf{153 \pm 1}$ & $\mathbf{999 \pm 5}$  & $\mathbf{465 \pm 2}$ \\
 & 50/50    & $\mathbf{803 \pm 35}$ & $\mathbf{1782 \pm 34}$ & $\mathbf{1539 \pm 38}$ \\
\bottomrule
\end{tabular}%
}
\end{wraptable}

\paragraph{Simulation Results.}~Table~\ref{tab:vla-quant-results} reports closed-loop success rate on LIBERO across the bit-width sweep. On OpenVLA-OFT, ActQuant retains $95.0\%$ at $3.0$\,bpw and $90.1\%$ at $2.5$\,bpw, against an fp16 ceiling of $96.9\%$. AWQ and GPTQ remain functional at $3.0$\,bpw ($91.5\%$ and $89.0\%$, both below ActQuant) but score $0.0\%$ on every suite at $2.0$\,bpw, while QVLA drops to $37.4\%$ already at $3.0$\,bpw and to $0.0\%$ at $2.5$\,bpw, the budget at which it is the only baseline we evaluate against ActQuant. On $\pi_{0.5}$, ActQuant achieves $94.8\%$ at $3.0$\,bpw and $85.7\%$ at $2.5$\,bpw against QVLA's $79.2\%$, and at $2.0$\,bpw it is the only evaluated method that still completes tasks ($48.0\%$), where AWQ, GPTQ, and QVLA all score $0.0\%$. At its lowest bit-width on OpenVLA-OFT, ActQuant compresses the model from $14.3$\,GB to $2.7$\,GB, a $5.3\times$ reduction.

\paragraph{Real-World Robot Deployment.}
%\label{sec:exp-realworld}
To verify that simulation gains transfer to physical hardware, we evaluate ActQuant on the UR3 arm described in Section~\ref{sec:exp-setup}. Across four manipulation tasks (Table~\ref{tab:realworld-results}), our $\pi_{0.5}$ deployment at $3.0$\,bpw retains $75.0\%$ average success rate against the $77.5\%$ fp16 baseline, while compressing the model from $6.7$\,GB to $2.7$\,GB ($2.5\times$). The success-vs-memory gains we observe in simulation therefore carry over to a real robot.

\paragraph{Long-Horizon Bimanual Manipulation.}
Single-arm pick-and-place exercises only a narrow slice of the behaviors a deployed policy must retain, so we additionally evaluate on a bimanual UR3 platform over three tasks, at $20$ trials per task (Table~\ref{tab:bimanual-results}). In \emph{place screwdriver}, the policy picks up a large black screwdriver and places it on a target space. In \emph{disassemble screwdriver}, it picks the orange screwdriver from the table, removes its tip with the second arm, places the tip inside a black box, and then picks up the screwdriver body and returns it. In \emph{screw hand-off}, it picks a screw on the left, hands it from the left arm to the second arm and then to the right arm, and places it in the screw box. The latter two are long-horizon dual-arm manipulations of a kind that current simulation benchmarks do not fully cover. Averaged over the three tasks, ActQuant reaches $53.3\%$ at $3.0$\,bpw and $50.0\%$ at $2.5$\,bpw, against $41.7\%$ for AWQ at $3.0$\,bpw and $58.3\%$ for the fp16 baseline. The gap is widest exactly where quantization bites hardest: on the two long-horizon dual-arm tasks, ActQuant at $3.0$\,bpw matches the fp16 policy on \emph{disassemble screwdriver} ($7/20$) and reaches $8/20$ on \emph{screw hand-off}, whereas AWQ falls to $5/20$ on both; even on the comparatively easy \emph{place screwdriver}, AWQ drops to $15/20$ while ActQuant holds $17/20$ against fp16's $18/20$. ActQuant outperforms AWQ on every task at the matched $3.0$\,bpw budget, and at $2.5$\,bpw it still meets or exceeds AWQ's $3.0$\,bpw counts on all three. The long-horizon dual-arm skills that a standard PTQ baseline loses below four bits are therefore preserved under action-guided allocation.

\paragraph{Inference Speedup.}~Table~\ref{tab:openvla-speed} reports per-token inference latency for OpenVLA across three hardware platforms, an NVIDIA RTX A6000 workstation, an Apple MacBook Pro M4 Pro, and an NVIDIA AGX Thor edge box. The Q4\_K\_M build produced by OmniModel.cpp consistently achieves the lowest latency, with up to $1.5\times$ speedup over the PyTorch reference on the RTX A6000 ($233\!\to\!153$\,ms), $1.3\times$ on the M4 Pro ($1319\!\to\!999$\,ms), and $1.3\times$ on the AGX Thor ($598\!\to\!465$\,ms). Unlike PyTorch, the C/C++ build can also fall back to a 50/50 CPU/GPU split when on-board VRAM is the limiting resource, making the same quantized model deployable on edge boxes that cannot fit the fp16 backbone in GPU memory.

\begin{table}[t]
\centering
\caption{Real-world success-rate and memory comparison on a $6$-DoF Universal Robots UR3 arm. }
\label{tab:realworld-results}
\resizebox{0.95\textwidth}{!}{%
\begin{tabular}{l| c |cccc| c| c}
\toprule
\multirow{2}{*}{\textbf{Method}} & \multirow{2}{*}{\shortstack{\textbf{Vision+LLM}\\\textbf{BPW}}} & \multicolumn{4}{c}{\textbf{Real-world Task}} & \multirow{2}{*}{\shortstack{\textbf{Average}\\\textbf{SR}}} & \multirow{2}{*}{\shortstack{\textbf{Mem.}\\\textbf{(GB)}}} 
% & \multirow{2}{*}{\textbf{Speedup}} 
\\
\cmidrule(lr){3-6}
 & & Pick pink cylinder & Pick white cup & Remove nested cups & Single-finger push & &  \\
\midrule
$\pi_{0.5}$        & 16.0 & 18/20 & 14/20 & 16/20 & 14/20 & 77.5\% & 6.7  \\
\rowcolor{gray!15}
\textbf{ActQuant}  & 3.0  & 14/20 & 14/20 & 16/20 & 16/20 & 75.0\% & 2.7   \\
\bottomrule
\end{tabular}%
}
\end{table}

\paragraph{Ablation Analysis.}      
\label{sec:exp-ablation}

\begin{table}[t]
\centering
\caption{Real-world evaluation on the bimanual UR3 platform; the last two tasks are long-horizon dual-arm manipulations. Each task is reported as successful trials out of $20$.}
\label{tab:bimanual-results}
\small
\begin{tabular}{@{}l c ccc c@{}}
\toprule
\multirow{2}{*}{\textbf{Method}} & \multirow{2}{*}{\shortstack{\textbf{Vision+LLM}\\\textbf{BPW}}} & \multicolumn{3}{c}{\textbf{Real-world Task}} & \multirow{2}{*}{\shortstack{\textbf{Average}\\\textbf{SR}}} \\
\cmidrule(lr){3-5}
 & & Place screwdriver & Disassemble screwdriver & Screw hand-off & \\
\midrule
$\pi_{0.5}$ & 16.0 & 18/20 & 7/20 & 10/20 & 58.3\% \\
AWQ         & 3.0  & 15/20 & 5/20 & 5/20  & 41.7\% \\
\rowcolor{gray!15}
\textbf{ActQuant} & 3.0 & \textbf{17/20} & \textbf{7/20} & \textbf{8/20} & \textbf{53.3\%} \\
\rowcolor{gray!15}
\textbf{ActQuant} & 2.5 & \textbf{17/20} & 6/20 & 7/20 & 50.0\% \\
\bottomrule
\end{tabular}
\end{table}

% Add to preamble if \diff is not already defined:
% \newcommand{\diff}[1]{\textcolor{gray}{#1}}

\begin{table}[h]
\centering
\caption{Ablation study of proposed components on OpenVLA-OFT  evaluated on LIBERO. }
\label{tab:ablation}
\resizebox{0.97\textwidth}{!}{%
\begin{tabular}{l c ccccc c c}
\toprule
\multirow{2}{*}{\textbf{Method}} & \textbf{Vision+LLM} & \multicolumn{5}{c}{\textbf{Success Rate (\%)}} & \multirow{2}{*}{\textbf{$\Delta$}} & \multirow{2}{*}{\textbf{Mem (GB)}} \\
\cmidrule(lr){3-7}
 & \textbf{BPW} & \textbf{Spatial} & \textbf{Object} & \textbf{Goal} & \textbf{Long} & \textbf{Average} & & \\
\midrule
\multicolumn{9}{l}{\textit{OpenVLA-OFT}} \\
\midrule
fp16 (baseline)                            & 16.0 & 97.6 & 98.4 & 96.8 & 95.1 & 96.9 & 0.00 & 14.3 \\
RTN block quantization                     & 2.6  & 0.0  & 17.8 & 0.0  & 0.0  & 4.5  & $-92.4$       & 2.7 \\
\quad + magnitude-weighted scale           & 2.6  & 69.4 & 89.6 & 19.0 & 52.6 & 57.7 & $-39.2$       & 2.7 \\
\quad\quad + Action-Only Fisher            & 2.6  & 87.6 & 98.4 & 77.2 & 90.4 & 88.4 & $-8.5$        & 2.7 \\
\quad\quad\quad + Action-Mixed Fisher      & 2.6  & 88.8 & 98.0 & 81.2 & 91.4 & 89.9 & $-7.0$        & 2.7 \\
\quad\quad\quad\quad + HSIC bit allocation & 2.7  & 92.0 & 98.4 & 85.8 & 92.2 & 92.1 & $-4.8$        & 2.7 \\
% \quad\quad\quad + Action head FT         & -- & -- & -- & -- & -- & -- & --          & -- \\
\bottomrule
\end{tabular}
}
\end{table}

Table~\ref{tab:ablation} isolates each ActQuant component on top of an RTN block-quantization baseline at $\sim$$2.6$ average bits-per-weight on OpenVLA-OFT. RTN alone collapses to $4.5\%$, near random. Switching the per-block scale objective from unweighted MSE to a magnitude-weighted MSE recovers the bulk of the gap ($57.7\%$), confirming that scale choice rather than codebook design dominates below four bits. Replacing weight magnitude with the Action-Only Fisher signal raises success rate to $88.4\%$, a $30.7\%$ gain attributable to action-aware sensitivities, and adding the LM-head pathway to form the full Action-Mixed Fisher (Eq.~\eqref{eq:amf-fisher}) lifts it further to $89.9\%$. Layering the HSIC-based inter-tensor bit allocation on top closes the remaining gap to $92.1\%$, only $4.8$ points below the fp16 reference. The two proposed components contribute independently and compose additively.

\section{Conclusion}

We presented \textbf{ActQuant}, an action-guided mixed-precision PTQ framework that pushes VLA models below four bits with an HSIC-based inter-tensor bit allocator that scores each weight matrix by its action relevance and an intra-tensor scale optimizer weighted by an Action-Mixed Fisher signal that focuses dynamic range on weights most influential for control, paired with \textbf{OmniModel.cpp}, an agentic pipeline that runs the quantized models in a native C/C++ runtime with efficient low-bit kernels. On LIBERO, ActQuant retains $95.0\%$ on OpenVLA-OFT and $94.8\%$ on $\pi_{0.5}$ at $3.0$\,bpw; at $2.5$\,bpw it compresses OpenVLA-OFT by $5.3\times$ at $90.1\%$ closed-loop success where QVLA scores $0.0\%$, and on $\pi_{0.5}$ it is the only evaluated method still functional at $2.0$\,bpw. On a 6-DoF UR3 arm it compresses $\pi_{0.5}$ by $2.5\times$ with almost no drop in success rate, and on a bimanual platform it preserves long-horizon dual-arm skills that AWQ loses at the same budget.

\section*{Acknowledgements}
The authors gratefully acknowledge support from the National Science Foundation through grant 2414652, and from EmbodyX for both financial support and providing the computational resources used in this work.

\bibliographystyle{plain}
\bibliography{refs}

% Arman: I added the newpage 
\newpage
%%%%%%%%%%%%%%%%%%%%%%%%%%%%%%%%%%%%%%%%%%%%%%%%%%%%%%%%%%%%

\appendix

\section{Hilbert-Schmidt Independence Criterion}
\label{app:hsic}

This appendix expands on the HSIC definition reviewed in the Background
section. For two random variables $A$ and $B$ with associated kernel
functions $k_A$ and $k_B$ on their respective domains, the
\emph{population-level} Hilbert-Schmidt Independence Criterion is
\begin{align}
\label{eq:hsic-pop}
\mathrm{HSIC}(A, B) \;&=\; \mathbb{E}_{A,B,A',B'}\!\left[k_A(A,A')\, k_B(B,B')\right] \notag \\
&+\; \mathbb{E}_{A,A'}\!\left[k_A(A,A')\right]\, \mathbb{E}_{B,B'}\!\left[k_B(B,B')\right] \notag \\
&-\; 2\,\mathbb{E}_{A,B}\!\left[\mathbb{E}_{A'}\!\left[k_A(A,A')\right]\, \mathbb{E}_{B'}\!\left[k_B(B,B')\right]\right],
\end{align}
where $A', B'$ are independent copies of $A, B$, respectively. HSIC is
the squared Hilbert-Schmidt norm of the cross-covariance operator
between feature maps in the two reproducing kernel Hilbert spaces, and
under \emph{universal} kernels (e.g., the Gaussian RBF kernel)
$\mathrm{HSIC}(A, B) = 0$ if and only if $A$ and $B$ are independent.
HSIC therefore acts as a faithful dependence measure that, in contrast
to mutual information, requires no density estimation.

\paragraph{Empirical estimator.}
Given $K$ i.i.d.\ samples $\{(a_k, b_k)\}_{k=1}^{K}$, the population
HSIC of Eq.~\eqref{eq:hsic-pop} is estimated by the trace expression
of Eq.~\eqref{eq:hsic-emp} in the main text, with $\mathbf{C} =
\mathbf{I} - \tfrac{1}{K}\mathbf{1}\mathbf{1}^{\top}$ centering the
kernel matrices in feature space. The estimator has bias
$\mathcal{O}(K^{-1})$ and converges at rate $\mathcal{O}(K^{-1/2})$
under standard regularity conditions, and its trace form lets it be
computed in $\mathcal{O}(K^{2})$ time per pair of variables, well
within the calibration budget of an action-aware PTQ pipeline.

\paragraph{Use as an information-theoretic surrogate.}
HSIC has been used as a tractable substitute for mutual information in
information-bottleneck objectives~\cite{ma2020hsic, hbar,wang2023dualhsic} and as a
feature-selection signal in deep models~\cite{li2021self, liu2022hsic, miklautz2025h}. We use this same property in
Section~\ref{sec:hsic} to build an per-tensor sensitivity score that
penalizes redundancy with the model input and rewards alignment with
the ground-truth actions, without ever differentiating through the
action head.

\section{Hessian-Weighted Reconstruction and Fisher Approximation}
\label{app:hessian}

\paragraph{Hessian-weighted reconstruction objective.}
We derive the Hessian-weighted objective referenced in
Eq.~\eqref{eq:quant-obj} of the Background section. Quantizing
$\boldsymbol{\theta}$ to $\boldsymbol{\hat{\theta}}$ introduces a
perturbation $\delta = \boldsymbol{\hat{\theta}} - \boldsymbol{\theta}$,
and the deployment objective is to minimize the resulting change in the
action loss,
\begin{equation}
\boldsymbol{\hat{\theta}}^{\star} \;=\; \operatorname*{\arg\min}_{\boldsymbol{\hat{\theta}}}\; \mathcal{L}(\boldsymbol{\hat{\theta}}) - \mathcal{L}(\boldsymbol{\theta}).
\end{equation}
A second-order Taylor expansion of $\mathcal{L}$ around $\boldsymbol{\theta}$ gives
\begin{equation}
\mathcal{L}(\boldsymbol{\hat{\theta}}) - \mathcal{L}(\boldsymbol{\theta}) \;\approx\; \bigl(\nabla_{\boldsymbol{\theta}}\mathcal{L}\bigr)^{\top} \delta + \tfrac{1}{2}\, \delta^{\top}\, \mathbf{H}\, \delta,
\end{equation}
with $\mathbf{H} = \nabla^{2}_{\boldsymbol{\theta}}\mathcal{L}$. The
calibration distribution coincides with the fine-tuning distribution, so
the backbone sits near a local minimum of $\mathcal{L}$ and
$\nabla_{\boldsymbol{\theta}}\mathcal{L} \approx \mathbf{0}$. The
first-order term is therefore negligible, and the deployment objective
reduces to the Hessian-weighted reconstruction in
Eq.~\eqref{eq:quant-obj}. The diagonal entry $H_{ii}$ measures the local
curvature of $\mathcal{L}$ along $\theta_i$ and is used as a per-element
sensitivity in both the inter-tensor allocator (Section~\ref{sec:hsic})
and the intra-tensor scale optimizer (Section~\ref{sec:scale-opt}).

\paragraph{Why we replace $\mathbf{H}$ by a surrogate.}
Materializing the full Hessian is intractable at backbone scale: $\mathbf{H} \in \mathbb{R}^{P \times P}$ has $\mathcal{O}(P^{2})$ entries, and forming any one entry requires a Hessian-vector product. For a 7B-parameter LLM backbone, $P \approx 7 \times 10^{9}$, so even storing the diagonal $\{H_{ii}\}_{i=1}^{P}$ is feasible only via gradient-style estimators. We therefore replace $\mathbf{H}$ in Eq.~\eqref{eq:quant-obj} by a tractable surrogate that captures its diagonal at a local minimum.

\paragraph{Fisher information and its equivalence to the Hessian at the optimum.}
Treating the action loss as a negative log-likelihood,
$\mathcal{L}(\boldsymbol{\theta}) \;=\; -\mathbb{E}_{(x, y) \sim \mathcal{D}}\!\left[\log p_{\boldsymbol{\theta}}(y \mid x)\right]$, the (true) Fisher information matrix is
\begin{equation}
\mathbf{F}(\boldsymbol{\theta}) \;=\; \mathbb{E}_{(x, y) \sim p_{\boldsymbol{\theta}}}\!\left[ \nabla_{\boldsymbol{\theta}} \log p_{\boldsymbol{\theta}}(y \mid x)\, \nabla_{\boldsymbol{\theta}} \log p_{\boldsymbol{\theta}}(y \mid x)^{\top} \right].
\end{equation}
A standard result in natural-gradient theory~\cite{martens2020new} establishes that $\mathbf{F}(\boldsymbol{\theta}) = \mathbf{H}(\boldsymbol{\theta})$ when the model distribution coincides with the data distribution; for a fine-tuned VLA, $\boldsymbol{\theta}$ sits sufficiently close to such a fixed point that $\mathbf{F} \approx \mathbf{H}$ holds throughout calibration.

\paragraph{Empirical Fisher and diagonal approximation.}
In practice the inner expectation is replaced by samples drawn from the calibration set $\mathcal{D}$, giving the empirical Fisher matrix of Eq.~\eqref{eq:fisher-bg}. To make scoring tractable per element we further replace the full $\mathbf{F}$ by its diagonal,
\begin{equation}
\mathcal{F}_{ii} \;=\; \frac{1}{|\mathcal{D}|} \sum_{d \in \mathcal{D}} \left( \frac{\partial \mathcal{L}_d}{\partial \theta_i} \right)^{2}, \qquad i = 1, \ldots, P,
\end{equation}
so that computing $\{\mathcal{F}_{ii}\}_{i=1}^{P}$ requires only first-order information: a single backward pass per calibration sample, $\mathcal{O}(P)$ memory in total, and no Hessian-vector products. We adopt the diagonal approximation $\mathbf{H} \approx \mathrm{diag}(\mathcal{F})$ throughout, which assumes that cross-parameter interactions are negligible relative to per-parameter curvature. The methodology refines this surrogate further by replacing $\mathcal{L}_d$ with the Action-Mixed Fisher loss (Section~\ref{sec:scale-opt}), so that $\mathcal{F}_{ii}$ reflects gradient signal from both the action-head and (when available) the categorical pathways.

\section{Cross-Pathway Covariance in AMF}
\label{app:cross-cov}

This appendix expands on the AMF Fisher diagonal of Eq.~\eqref{eq:amf-fisher} when the architecture exposes both an action-head loss $\mathcal{L}^{\mathrm{act}}_d$ and an LM-head categorical loss $\mathcal{L}^{\mathrm{cls}}_d$ (e.g., OpenVLA-OFT). Architectures that lack a categorical head reduce to the $\alpha = 1$ case below and the rest of this section is moot. Let
\[
\nabla_i\mathcal{L}^{\mathrm{act}}_d \;\equiv\; \frac{\partial\mathcal{L}^{\mathrm{act}}_d}{\partial\theta_i},
\qquad
\nabla_i\mathcal{L}^{\mathrm{cls}}_d \;\equiv\; \frac{\partial\mathcal{L}^{\mathrm{cls}}_d}{\partial\theta_i}.
\]
By linearity of differentiation and the definition of $\mathcal{L}^{\mathrm{AMF}}_d$ in Eq.~\eqref{eq:amf-loss},
\[
\frac{\partial\mathcal{L}^{\mathrm{AMF}}_d}{\partial\theta_i} \;=\; \alpha\,\nabla_i\mathcal{L}^{\mathrm{act}}_d \;+\; (1-\alpha)\,\nabla_i\mathcal{L}^{\mathrm{cls}}_d.
\]
Substituting into Eq.~\eqref{eq:amf-fisher} and expanding the square gives the decomposition
\begin{equation}
\label{eq:amf-decomposition}
\mathcal{F}^{\mathrm{AMF}}_{ii} \;=\; \alpha^{2}\,\mathcal{F}^{\mathrm{act}}_{ii} \;+\; (1-\alpha)^{2}\,\mathcal{F}^{\mathrm{cls}}_{ii} \;+\; 2\alpha(1-\alpha)\,\mathcal{C}_{ii},
\end{equation}
where the per-pathway Fisher diagonals and the cross-pathway covariance are
\[
\mathcal{F}^{\mathrm{act}}_{ii} = \frac{1}{|\mathcal{D}|}\sum_{d}\bigl(\nabla_i\mathcal{L}^{\mathrm{act}}_d\bigr)^{2},
\quad
\mathcal{F}^{\mathrm{cls}}_{ii} = \frac{1}{|\mathcal{D}|}\sum_{d}\bigl(\nabla_i\mathcal{L}^{\mathrm{cls}}_d\bigr)^{2},
\quad
\mathcal{C}_{ii} = \frac{1}{|\mathcal{D}|}\sum_{d}\nabla_i\mathcal{L}^{\mathrm{act}}_d\,\nabla_i\mathcal{L}^{\mathrm{cls}}_d.
\]

The covariance term $\mathcal{C}_{ii}$ vanishes identically for either pure single-loss Fisher ($\alpha \in \{0, 1\}$) and is therefore unique to the blended formulation. Its sign at parameter $\theta_i$ encodes whether perturbations of $\theta_i$ shift the action-head and language-modeling outputs in correlated directions: positive $\mathcal{C}_{ii}$ identifies parameters jointly load-bearing for both heads (typical of weights that mediate shared task semantics) and amplifies their AMF importance, while negative $\mathcal{C}_{ii}$ down-weights parameters whose pathways pull in opposite directions, which would otherwise be assigned spuriously high importance under a single-loss Fisher.

The weighting prefactor $2\alpha(1-\alpha)$ in Eq.~\eqref{eq:amf-decomposition} is maximized at $\alpha = 1/2$, where the cross-covariance contribution to $\mathcal{F}^{\mathrm{AMF}}_{ii}$ is largest in absolute terms. This motivates our default choice of $\alpha = 1/2$ for OpenVLA-OFT in the experiments of Section~\ref{sec:experiments}.

\section{OmniModel.cpp: Agentic Pipeline for VLA Deployment}
\label{app:omnimodel}

Native C/C++ inference is what actually delivers the on-device benefits of extreme quantization: it removes the Python/PyTorch~\cite{paszke2019pytorch} runtime and its dependencies, and gives the model access to hand-written kernels for packed sub-4-bit weight formats and to flexible CPU/GPU memory offloading. The de facto standard runtime for this on edge hardware is the GGML tensor library~\cite{ggml} together with \texttt{llama.cpp}~\cite{llamacpp}, which stores quantized weights in an on-disk binary container called \emph{GGUF}. A GGUF file holds, for every weight tensor, the packed integer codes together with the per-block scales, zero points, and the quantization type tag (e.g., Q4\_K\_M, Q3\_K\_S) needed to dequantize on the fly inside a hand-written GGML kernel. ActQuant's per-block scale and per-tensor bit-width outputs map directly onto this format, so any backbone whose forward graph is expressed in GGML can be deployed end-to-end in C/C++.

The catch is that \texttt{llama.cpp} natively supports only a narrow set of decoder-only LLM architectures, with no built-in support for VLA backbones such as OpenVLA-OFT (a SigLIP~\cite{zhai2023siglip}+DINOv2~\cite{oquab2024dinov2} vision tower over a LLaMA~2~\cite{touvron2023llama} stack with an L1-regression action head) or $\pi_{0.5}$ (a vision-language stack with a flow-matching action head). Porting a new VLA backbone into the GGML computation graph traditionally requires weeks of by-hand engineering: reconstructing the forward graph module by module, serializing each parameter tensor into the GGUF container, implementing custom GGML kernels for operators without a direct equivalent, and debugging numerical drift between the PyTorch reference and the C++ implementation. To solve this issue and be able to deploy various quantized models beyond LLMs efficiently on edge, we propose \textbf{OmniModel.cpp}, an agentic conversion pipeline used to obtain every C/C++ deployment artifact in this paper.

\begin{figure}[t]
    \centering
    \includegraphics[width=\linewidth]{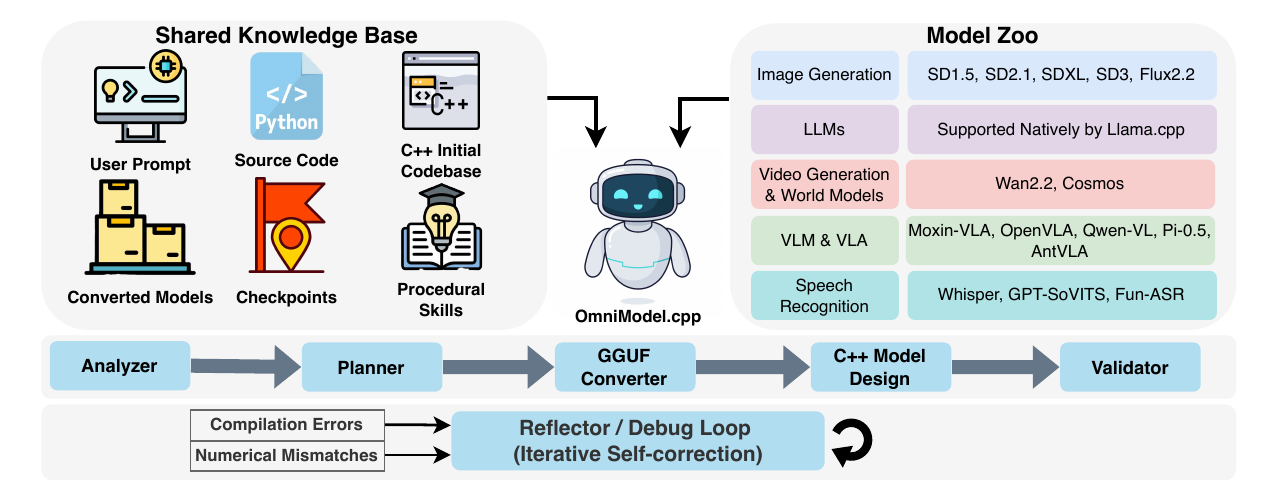}
    \caption{OmniModel.cpp pipeline for deploying ActQuant-quantized VLA backbones in native C/C++. A shared knowledge base and a library of procedural skills feed a planner-driven loop (analyze, plan, convert to GGUF, build the GGML graph, validate); a reflector iterates on compilation errors and per-tensor numerical mismatches against the PyTorch reference until parity is reached.}
    \label{fig:omnimodel-pipeline}
\end{figure}

OmniModel.cpp organizes a code-generation agent into three cooperating layers, illustrated in Figure~\ref{fig:omnimodel-pipeline}. (i)~A \emph{shared knowledge base} stores the PyTorch source and checkpoints, an initial C++ skeleton, and a repository of previously converted backbones, so that a new conversion can reuse validated solutions rather than reasoning from scratch. (ii)~A library of \emph{dynamic procedural skills} encodes self-contained, reusable subtasks (serializing a particular submodule to GGUF, registering a custom GGML operator, wiring up a Python binding) and is updated whenever a new architecture-specific pattern is resolved. (iii)~An \emph{automated verification loop} runs the agent inside a sandbox with file I/O, shell-driven compilation, and semantic codebase search; the agent caches per-module intermediate tensors from the PyTorch reference, and after each module is compiled it tests cosine similarity against this cache. When similarity falls below $\tau = 0.99$, the reflector isolates the failing tensor, diagnoses the root cause, patches the implementation, and rebuilds, iterating until numerical parity is achieved. The conversion itself proceeds through a fixed sequence (analyzer, planner, GGUF converter, C++ model design, validator), with the GGUF converter serializing the ActQuant per-block scales and per-tensor bit-widths produced by our quantization algorithm.

\section{Additional Information on Calibration Data for LIBERO Simulation}
\label{app:calibration}

We use the jointly fine-tuned OpenVLA-OFT checkpoint that has been fine-tuned on all four LIBERO task suites (Spatial, Object, Goal, and Long), and draw $60$ episodes from the LIBERO fine-tuning split as our calibration set. Prior work such as QVLA~\cite{qvla} relies on $512$ episodes (roughly one third of the entire fine-tuning split), which is impractical for post-training quantization. Our focus is on preserving action-generation quality after quantization with an adequate, not maximal, calibration budget. The $60$ episodes are not drawn at random: each LIBERO suite contains $10$ task descriptions, and we sample one episode per task description from Spatial, Object, and Goal; for LIBERO-Long, whose tasks are long-horizon and more complex, we sample three episodes per task description, yielding the minimum of $60$ episodes total. The episode-level random sampling within each task description uses a fixed random seed of $42$ for reproducibility.

The same $60$-episode calibration set is reused to compute both the HSIC-based per-tensor sensitivity scores (Eq.~\eqref{eq:hsic-sens}) and the Action-Mixed Fisher per-element importance metric (Eq.~\eqref{eq:amf-fisher}), so the two ActQuant signals are derived from a single shared calibration pass.

\section{Robustness to the Calibration Set}
\label{app:calib-robust}

ActQuant derives both of its signals from a single $60$-episode calibration set (Appendix~\ref{app:calibration}), which raises the question of how much the result depends on that particular draw. We test this along two axes, the random seed used to draw the episodes and the size of the set, on the $\pi_{0.5}$ backbone at the $2.5$\,bpw operating point of Table~\ref{tab:vla-quant-results}.

\paragraph{Calibration seed.}
We draw the $60$-episode set ($10/10/10/30$ across Spatial, Object, Goal, and Long) under four seeds: the one used throughout the paper and three additional ones. For each seed we rerun the entire pipeline independently, computing the AMF Fisher, solving the HSIC allocation, and quantizing, and then evaluate closed-loop on all four suites at $20$ trials per task ($800$ rollouts per seed). Table~\ref{tab:calib-seed} reports the outcome.

\begin{table}[h]
\centering
\caption{Closed-loop success rate (\%) on LIBERO for $\pi_{0.5}$ at $2.5$\,bpw under four independent calibration draws. The first row is the entry reported in Table~\ref{tab:vla-quant-results}.}
\label{tab:calib-seed}
\small
\begin{tabular}{@{}l cccc c@{}}
\toprule
\textbf{Calibration seed} & \textbf{Spatial} & \textbf{Object} & \textbf{Goal} & \textbf{Long} & \textbf{Avg.} \\
\midrule
Paper seed (Table~\ref{tab:vla-quant-results}) & 96.6 & 98.4 & 74.0 & 73.6 & 85.7 \\
Seed 43 & 95.5 & 96.5 & 77.0 & 75.0 & 86.0 \\
Seed 44 & 93.0 & 96.0 & 78.0 & 79.0 & 86.5 \\
Seed 45 & 93.0 & 96.5 & 85.0 & 81.0 & 88.9 \\
\bottomrule
\end{tabular}
\end{table}

Across the four draws the average success rate is $86.8 \pm 1.5$ (range $85.7$ to $88.9$). The spread is small, and it is not a degradation: the draw that deviates most, seed $45$, does so upward, so no calibration set we tried harms the quantized policy relative to the one reported in the main text. The allocation is equally stable. Reading the per-tensor quantization types off the four quantized backbones, $95.2\%$ of the $126$ LLM weight tensors receive an identical type under all four seeds, and each of the six tensors that differ moves by at most $0.375$\,bpw, one or two adjacent steps in the type menu. Calibration randomness therefore perturbs neither the bit assignment nor the downstream success rate in a way that costs performance.

\paragraph{Calibration set size.}
We also vary the amount of calibration data, holding the seed fixed and rebuilding the full pipeline from a $40$-episode set ($10/10/10/10$) in place of the $60$-episode set ($10/10/10/30$).

\begin{table}[h]
\centering
\caption{Effect of calibration set size on $\pi_{0.5}$ at $2.5$\,bpw, with the draw seed held fixed.}
\label{tab:calib-size}
\small
\begin{tabular}{@{}l cccc c@{}}
\toprule
\textbf{Calibration set} & \textbf{Spatial} & \textbf{Object} & \textbf{Goal} & \textbf{Long} & \textbf{Avg.} \\
\midrule
$60$ episodes (paper seed) & 96.6 & 98.4 & 74.0 & 73.6 & 85.7 \\
$40$ episodes (same seed)  & 92.0 & 96.5 & 75.5 & 73.5 & 84.4 \\
\bottomrule
\end{tabular}
\end{table}

Cutting the calibration set by a third costs $1.3$ points of average success rate ($85.7\%$ to $84.4\%$). The long-horizon suite is essentially unchanged ($73.6\%$ to $73.5\%$) even though the $40$-episode set allots it only a third as many long-horizon episodes ($10$ rather than $30$). Taken together with the seed sweep, this indicates that ActQuant is robust to both the composition and the size of the calibration set, which matters because the method is designed for the regime in which calibration data is scarce (Appendix~\ref{app:qvla-repro}).

\section{Hyperparameter Settings and Sensitivity}
\label{app:hyperparams}

ActQuant exposes three hyperparameters: the two HSIC balancing weights $\lambda_{\mathrm{red}}$ and $\lambda_{\mathrm{act}}$ of the inter-tensor sensitivity score (Eq.~\eqref{eq:hsic-sens}), and the AMF mixing weight $\alpha$ of the intra-tensor objective (Eq.~\eqref{eq:amf-loss}). We use $\lambda_{\mathrm{red}} = \lambda_{\mathrm{act}} = 1$ throughout, and $\alpha = 0.5$ for OpenVLA-OFT and $\alpha = 1.0$ for $\pi_{0.5}$; the latter follows directly from the architecture, since $\pi_{0.5}$ exposes no action-token head and AMF therefore reduces to the Action-Only Fisher. This appendix reports how sensitive the method is to these choices. All sweeps use OpenVLA-OFT on LIBERO with $2{,}000$ closed-loop rollouts per point, under the calibration protocol of Appendix~\ref{app:calibration}.

\paragraph{AMF mixing weight.}
Table~\ref{tab:amf-sweep} sweeps $\alpha$ from the pure LM-head Fisher ($\alpha = 0$) to the pure Action-Only Fisher ($\alpha = 1$) at two bit-widths. The method is robust to this choice: the full range spans at most one point of average success rate, which is within run-to-run variance ($1\sigma \approx 0.5\%$ over $2{,}000$ episodes). Mixing the two pathways is nonetheless beneficial, most visibly at the tighter budget: at $3.0$\,bpw every mixed setting ($\alpha < 1$) outperforms the Action-Only Fisher, while at $3.5$\,bpw the settings are statistically indistinguishable. Our default $\alpha = 0.5$ sits in the top tier at both bit-widths and is additionally the value that maximizes the cross-pathway covariance prefactor $2\alpha(1-\alpha)$ of Eq.~\eqref{eq:amf-decomposition}. The sweep consists of independent re-runs, so the $\alpha = 0.5$ entry differs from the $95.0\%$ reported in Table~\ref{tab:vla-quant-results} by less than the run-to-run variance quoted above.

\begin{table}[h]
\centering
\caption{Sensitivity of ActQuant to the AMF mixing weight $\alpha$ (Eq.~\eqref{eq:amf-loss}) on OpenVLA-OFT, averaged over the four LIBERO suites ($2{,}000$ rollouts per entry). $\alpha = 0$ uses the LM-head Fisher alone and $\alpha = 1$ the Action-Only Fisher.}
\label{tab:amf-sweep}
\small
\begin{tabular}{@{}l c c@{}}
\toprule
\multirow{2}{*}{\textbf{AMF }$\boldsymbol{\alpha}$} & \multicolumn{2}{c}{\textbf{Average Success Rate (\%)}} \\
\cmidrule(lr){2-3}
 & \textbf{3.0\,bpw} & \textbf{3.5\,bpw} \\
\midrule
$0.0$ (LM-head only)        & 96.0 & 96.0 \\
$0.3$                       & 95.3 & 95.9 \\
$0.4$                       & 95.3 & 95.9 \\
$0.5$ (default)             & 95.4 & 95.7 \\
$0.7$                       & 95.2 & 96.2 \\
$1.0$ (Action-Only Fisher)  & 95.1 & 96.0 \\
\bottomrule
\end{tabular}
\end{table}

\paragraph{HSIC balancing weights.}
Table~\ref{tab:hsic-sweep} varies $\lambda_{\mathrm{red}}$ and $\lambda_{\mathrm{act}}$ by a factor of two in each direction around the default and re-runs the allocator of Eq.~\eqref{eq:hsic-alloc} at $3.0$\,bpw over the $224$ block tensors of the OpenVLA-OFT backbone. Scaling either weight leaves the bit allocation bit-for-bit identical, and the closed-loop success rate therefore unchanged, which confirms $\lambda_{\mathrm{red}} = \lambda_{\mathrm{act}} = 1$ as a robust default rather than a tuned choice. The invariance arises because the greedy allocator depends only on the induced ranking of per-layer errors, and on this calibration set the task-relevance term $\widehat{\mathrm{HSIC}}(Z, Y)$ is the dominant driver of that ranking, consistent with the action-guided design, with the input-redundancy term $\widehat{\mathrm{HSIC}}(X, Z)$ acting as the second half of the information-bottleneck objective rather than as a competing signal.

\begin{table}[h]
\centering
\caption{Sensitivity of the inter-tensor allocation to the HSIC balancing weights (Eq.~\eqref{eq:hsic-sens}) on OpenVLA-OFT at $3.0$\,bpw. ``Tensors reallocated'' counts the block tensors whose assigned bit-width changes relative to the default $\lambda_{\mathrm{red}} = \lambda_{\mathrm{act}} = 1$.}
\label{tab:hsic-sweep}
\small
\begin{tabular}{@{}c c c c@{}}
\toprule
$\boldsymbol{\lambda_{\mathrm{red}}}$ & $\boldsymbol{\lambda_{\mathrm{act}}}$ & \textbf{Tensors reallocated} & \textbf{Average SR (\%)} \\
\midrule
$0.5$ & $1$   & $0 / 224$ & 95.4 \\
$2$   & $1$   & $0 / 224$ & 95.4 \\
$1$   & $0.5$ & $0 / 224$ & 95.4 \\
$1$   & $2$   & $0 / 224$ & 95.4 \\
\bottomrule
\end{tabular}
\end{table}

\section{Per-Layer Sensitivity and Bit Allocation}
\label{app:alloc-viz}

\begin{figure}[h]
\centering
\includegraphics[width=\textwidth]{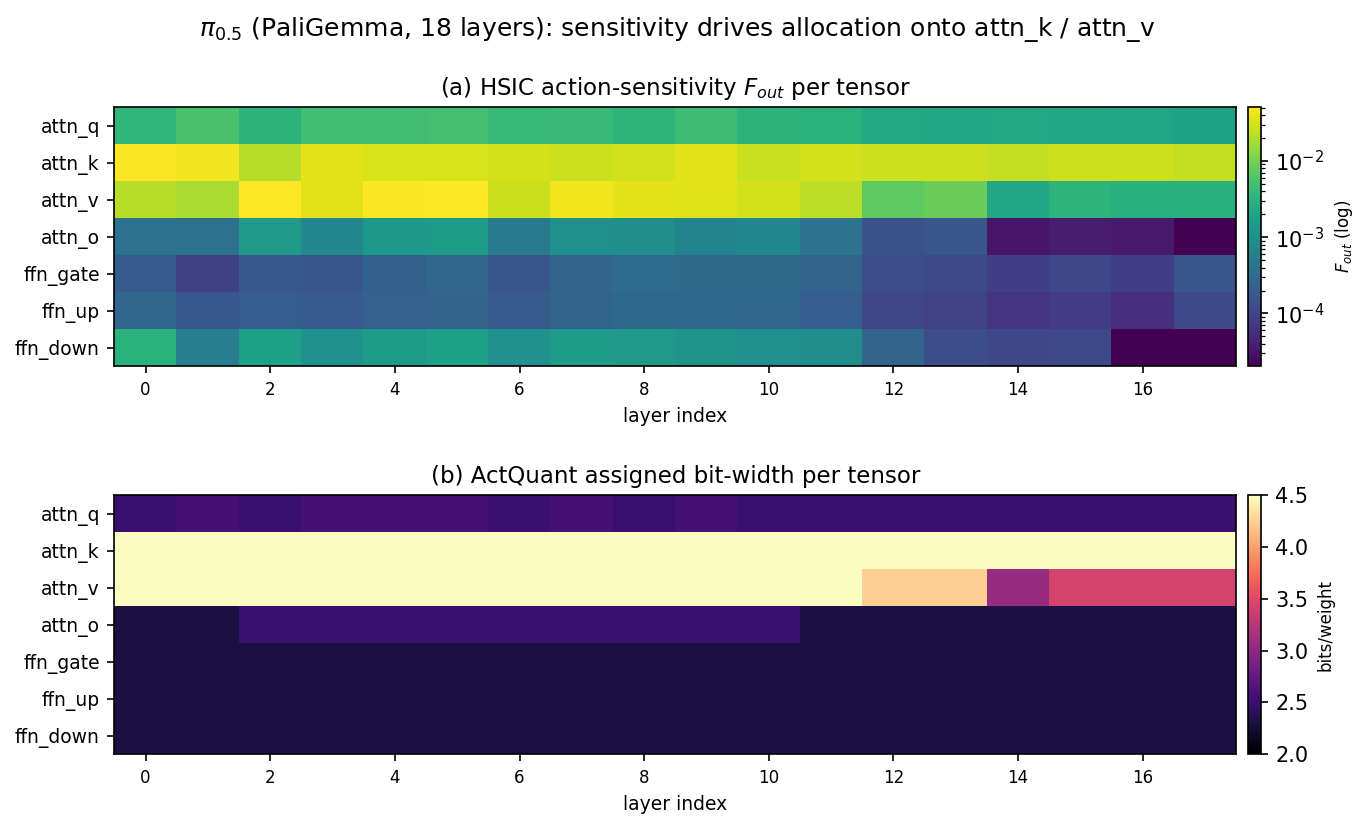}
\caption{Per-tensor HSIC action-sensitivity (a) and the resulting ActQuant bit-width assignment (b) across the $\pi_{0.5}$ language backbone, arranged by component and layer index.}
\label{fig:hsic-heatmap}
\end{figure}

To make the mechanics of Stage~1 concrete, Figure~\ref{fig:hsic-heatmap} visualizes both the input and the output of the inter-tensor allocator across every quantized tensor of the $\pi_{0.5}$ language backbone: panel~(a) shows the per-tensor HSIC action-sensitivity $\mathrm{S}(W_j^{(\ell)})$ of Eq.~\eqref{eq:hsic-sens}, and panel~(b) the bit-width the allocator subsequently assigns, both laid out as component $\times$ layer over the $18$ layers and $7$ component types ($126$ tensors) of the PaliGemma backbone.

Three observations follow from the two panels.

\paragraph{Allocated precision tracks sensitivity.}
The two panels are visibly aligned, and the agreement is quantitative: across the $126$ tensors, the per-tensor correlation between HSIC sensitivity and assigned bit-width is Pearson $r = 0.92$ and Spearman $\rho = 0.90$ ($n = 126$, $p < 10^{-15}$). The allocator does not scatter precision across the backbone; it concentrates it on the tensors the score flags as action-load-bearing. The sensitivity score, rather than the search procedure, is therefore what determines where the bits go, consistent with the matched-budget comparison of Appendix~\ref{app:alloc-designs}.

\paragraph{The allocation is component-driven rather than depth-driven.}
Precision separates far more sharply across component types than across depth. The key and value projections are raised to the top of the menu at nearly every layer (averaging roughly $4.5$ and $4.2$ bits per weight, with the value projections stepping down only in the last few layers), while the query and output projections and all three FFN matrices stay near the $\sim$$2.3$-bit floor. This is a mechanistically sensible reading of PaliGemma's grouped-query attention, which pairs eight query heads with a single key/value head: the KV projections are simultaneously the most action-load-bearing and, being far smaller than the query, output, and FFN matrices, the cheapest to protect at high precision under a fixed average-bits budget.

\paragraph{The action-relevance term drives the score.}
Decomposing Eq.~\eqref{eq:hsic-sens} into its two terms, the task-relevance term $\widehat{\mathrm{HSIC}}(Z, Y)$ exceeds the input-redundancy term $\widehat{\mathrm{HSIC}}(X, Z)$ at nearly every tensor. The ranking in panel~(a) is thus driven by how strongly a tensor's output aligns with the ground-truth actions, which is what the action-guided design intends, with the redundancy term acting as a correction rather than as the dominant signal. This is also consistent with the insensitivity of the allocation to the relative weighting of the two terms reported in Appendix~\ref{app:hyperparams}.

\section{Alternative Inter-Tensor Allocation Designs}
\label{app:alloc-designs}

Section~\ref{sec:hsic} allocates bit-widths from an HSIC sensitivity score, but several cheaper or more conventional allocators could fill the same slot. This appendix isolates that design choice and then examines what the resulting allocation actually selects.

\paragraph{Setup.}
We run the comparison on the $\pi_{0.5}$ backbone at the $2.5$\,bpw operating point of Table~\ref{tab:vla-quant-results}, a combined vision$+$LLM budget under which the language backbone averages $2.35$\,bpw over its $126$ linear layers and the vision tower is held at Q3\_K. The AMF per-element importance scores of Section~\ref{sec:scale-opt}, the budget, and the menu of admissible quantization types are all held fixed, so the only variable is the Stage-1 inter-tensor allocation. We compare three allocators: the built-in mixed-precision heuristic of \texttt{llama.cpp}, which assigns types per tensor from fixed transformer structure; a per-linear action-sensitivity proxy, which ranks tensors by a scalar action-sensitivity score derived from the same calibration pass and is a simpler action-aware signal than our criterion; and the HSIC allocator of Eq.~\eqref{eq:hsic-alloc}.

\paragraph{Open-loop action error.}
An allocation change perturbs the policy only mildly at any single step, so we first measure it with the most directly attributable metric available: the mean squared deviation of the quantized policy's action from the fp16 policy's action on held-out LIBERO calibration frames (Table~\ref{tab:alloc-mse}). HSIC lowers the per-step action error by $31\%$ relative to the \texttt{llama.cpp} heuristic and by $21\%$ relative to the per-linear proxy, at an identical budget and with an identical Stage-2 signal.

\begin{table}[h]
\centering
\caption{Per-step action deviation from the fp16 policy under three inter-tensor allocators, with the Stage-2 AMF signal and the budget held fixed.}
\label{tab:alloc-mse}
\small
\begin{tabular}{@{}l c@{}}
\toprule
\textbf{Allocation @ $2.5$\,bpw} & \textbf{Action MSE vs.\ fp16} ($\downarrow$) \\
\midrule
\texttt{llama.cpp} built-in mixed heuristic & 0.01271 \\
Per-linear action-sensitivity proxy         & 0.01110 \\
\textbf{HSIC (ours)}                        & \textbf{0.00879} \\
\bottomrule
\end{tabular}
\end{table}

\paragraph{Closed-loop success.}
The same ordering holds end-to-end. Table~\ref{tab:alloc-closed-loop} compares HSIC against the built-in heuristic over $800$ trials per configuration, where the heuristic reduces to a uniform assignment at this budget. HSIC gains $2.9$ points on average, and the gain is concentrated on the long-horizon suite ($74.5\%$ versus $68.0\%$), exactly where a poor allocation compounds most over a rollout (Appendix~\ref{app:compounding}); on the near-saturated Spatial and Object suites the two are within a few points. Because the closed-loop metric sits near its own ceiling at this budget, it is less sensitive to the allocation than the open-loop action error above, which is where the signal separates most cleanly. This sweep uses $800$ trials per configuration rather than the $2{,}000$ used for Table~\ref{tab:vla-quant-results}, so the HSIC row differs slightly from the corresponding ActQuant entry there.

\begin{table}[h]
\centering
\caption{Closed-loop success rate (\%) on LIBERO under two inter-tensor allocators, $800$ trials per configuration.}
\label{tab:alloc-closed-loop}
\small
\begin{tabular}{@{}l cccc c@{}}
\toprule
\textbf{Allocation @ $2.5$\,bpw} & \textbf{Spatial} & \textbf{Object} & \textbf{Goal} & \textbf{Long} & \textbf{Avg.} \\
\midrule
\texttt{llama.cpp} heuristic (uniform at this budget) & 93.0 & 95.5 & \textbf{76.0} & 68.0 & 83.1 \\
\textbf{HSIC (ours)} & \textbf{96.0} & \textbf{98.0} & 75.5 & \textbf{74.5} & \textbf{86.0} \\
\bottomrule
\end{tabular}
\end{table}

\paragraph{What the HSIC allocation selects.}
Beyond success rate, it is informative to ask whether the HSIC score recovers any structure known to matter in a transformer. The \texttt{llama.cpp} heuristic is effectively a residual-flow rule: its upcasts target the tensors that write directly into the residual stream, namely the value and output projections and the FFN down-projection, chosen from fixed architecture rather than from the task. The HSIC allocation both agrees and diverges with that expectation. It agrees on the value pathway, upcasting all $18$ value projections ($13$ of them to Q4\_K) and roughly half the output projections, so it recovers the residual-writer structure without being told about it. It diverges by additionally driving every key projection to Q4\_K and raising the query projections, which do not write into the residual stream but form the grouped-query-attention bottleneck the policy attends through, and by leaving the entire FFN at the $2$-bit base. We read the agreement as a sanity check rather than the validation target, since a criterion that merely reproduced the residual-flow rule could not improve on allocators built from it; the matched-budget gains in Tables~\ref{tab:alloc-mse} and~\ref{tab:alloc-closed-loop} come from the divergence toward the action-load-bearing attention pathway.

\paragraph{Relative contribution of the two stages.}
The Stage-2 AMF signal remains the larger of the two effects. Table~\ref{tab:ablation} replaces it with a magnitude-weighted proxy, a natural simpler design, and closed-loop success drops by $30.7$ points at a fixed bit-width ($57.7\%$ versus $88.4\%$ for the Action-Only Fisher, with the full Action-Mixed Fisher higher still), so the action-guided per-element signal does substantial work on its own. The inter-tensor allocator then contributes the remaining few points, and because closed-loop errors compound over a rollout (Appendix~\ref{app:compounding}), the per-step action gaps measured in Table~\ref{tab:alloc-mse} translate into the larger success-rate gaps of Table~\ref{tab:vla-quant-results}.

\section{Evaluation on RoboTwin 2.0}
\label{app:robotwin}

To test whether the results of Section~\ref{sec:exp-results} are specific to LIBERO, we additionally evaluate on RoboTwin 2.0, a simulator that differs from LIBERO in both embodiment and difficulty. RoboTwin 2.0 uses a bimanual dual-arm ALOHA-AgileX platform with a $14$-DoF action space and three camera views, and provides $50$ manipulation tasks. We run every task under both the clean and the domain-randomized setting at $20$ trials per task, giving $1{,}000$ closed-loop rollouts per setting.

We quantize the full policy, comprising the SigLIP vision tower and the PaliGemma language backbone, and compare against AWQ and GPTQ at matched $3$-bit and $2$-bit operating points, with the vision tower and the language backbone both quantized and the budget measured as the combined bits-per-weight over the quantized matmul weights. All three methods are calibrated on the same data, two demonstration episodes per task across all $50$ tasks ($100$ episodes in total), so the comparison isolates the quantization and allocation method rather than the calibration budget. Table~\ref{tab:robotwin} reports the outcome.

\begin{table}[h]
\centering
\caption{Closed-loop success rate on RoboTwin 2.0 under the clean and domain-randomized settings; the final column averages the two.}
\label{tab:robotwin}
\small
\begin{tabular}{@{}l c c c c@{}}
\toprule
\textbf{Method} & \textbf{BPW} & \textbf{Clean SR (\%)} & \textbf{Randomized SR (\%)} & \textbf{Average SR (\%)} \\
\midrule
fp16 (baseline) & 16.0 & 46.9 & 51.1 & 49.0 \\
\midrule
\multirow{2}{*}{AWQ}  & 3.0 & 37.4 & 37.7 & 37.6 \\
                      & 2.0 & 12.3 & 9.5  & 10.9 \\
\midrule
\multirow{2}{*}{GPTQ} & 3.0 & 39.5 & 41.2 & 40.4 \\
                      & 2.0 & 0.2  & 0.5  & 0.4  \\
\midrule
\multirow{2}{*}{\textbf{ActQuant}} & 3.0 & \textbf{40.0} & \textbf{42.5} & \textbf{41.3} \\
                                   & 2.0 & \textbf{41.2} & \textbf{41.9} & \textbf{41.6} \\
\bottomrule
\end{tabular}
\end{table}

At $3$\,bpw ActQuant is the strongest of the three methods under both settings: $42.5\%$ randomized against GPTQ's $41.2\%$ and AWQ's $37.7\%$, a paired gain of $4.8$ points over AWQ, and $40.0\%$ clean against $39.5\%$ and $37.4\%$. The separation widens sharply at $2$\,bpw, where both baselines break down, AWQ to $9.5\%$ randomized and $12.3\%$ clean and GPTQ to $0.5\%$ and $0.2\%$, while ActQuant holds near $41\%$ ($41.9\%$ randomized, $41.2\%$ clean), statistically indistinguishable from its own $3$-bit operating point. Averaged over the two settings, ActQuant scores $41.3\%$ at $3$\,bpw and $41.6\%$ at $2$\,bpw, against $40.4\%$ and $0.4\%$ for GPTQ and $37.6\%$ and $10.9\%$ for AWQ. This mirrors the LIBERO behavior reported in Table~\ref{tab:vla-quant-results}: below three bits the standard PTQ baselines fall to near-zero success while action-guided allocation keeps the policy functional, and the effect holds on a different embodiment, a larger action space, and under domain randomization.

\section{Hardware, Quantization Cost, and End-to-End Memory}
\label{app:hardware}

All experiments in this paper are run on three NVIDIA GPU classes. The fp16 baselines and the LIBERO closed-loop simulation rollouts use NVIDIA L40S GPUs ($48$\,GB GDDR6 VRAM each). The real-world UR3 deployment fine-tunes $\pi_{0.5}$ on NVIDIA A100 GPUs ($80$\,GB HBM2e) and serves the quantized policy at inference time on a single NVIDIA RTX A6000 ($48$\,GB GDDR6). One full ActQuant pass (HSIC-based inter-tensor allocation followed by Action-Mixed Fisher intra-tensor scale optimization on the $60$-episode calibration set) takes approximately two hours of wall-clock time per backbone on a single L40S, with peak GPU memory below the on-card $48$\,GB during both stages.

\paragraph{Weight memory versus end-to-end peak VRAM.}
The memory figures reported in Table~\ref{tab:vla-quant-results} and Table~\ref{tab:ablation} are full-model \emph{weight} memory: they cover the vision encoder, the language backbone, and the fp16 projector and action head, the latter two contributing roughly $0.4$\,GB that is held constant across bit-widths because those modules are never quantized. Weight memory is the quantity the allocator controls, but deployment on a fixed-VRAM edge device is governed by the peak resident footprint of the whole inference process. We therefore also measured peak resident VRAM during closed-loop LIBERO inference with OpenVLA-OFT, on a single $48$\,GB GPU at batch size $1$ through OmniModel.cpp; Table~\ref{tab:memory-breakdown} reports both quantities side by side.

\begin{table}[h]
\centering
\caption{Weight memory and end-to-end peak resident VRAM for OpenVLA-OFT during closed-loop LIBERO inference (batch size $1$, OmniModel.cpp, single $48$\,GB GPU). Weight memory includes the fp16 projector and action head, which are not quantized.}
\label{tab:memory-breakdown}
\small
\begin{tabular}{@{}l c c c@{}}
\toprule
\textbf{Configuration} & \textbf{Weight memory (GB)} & \textbf{Peak VRAM (GB)} & \textbf{Overhead (GB)} \\
\midrule
fp16                   & 14.3 & 15.3 & 1.0 \\
\midrule
ActQuant $4.0$\,bpw    & 4.2  & 5.3  & 1.1 \\
ActQuant $3.5$\,bpw    & 3.4  & 4.5  & 1.1 \\
ActQuant $3.0$\,bpw    & 3.2  & 4.3  & 1.1 \\
ActQuant $2.5$\,bpw    & 2.7  & 3.8  & 1.1 \\
\bottomrule
\end{tabular}
\end{table}

The overhead beyond weights is small and essentially constant at $\sim$$1.1$\,GB across every configuration: at batch-size-$1$ closed-loop control the KV cache and intermediate activations are negligible next to the weights, which is precisely the regime in which weight-only quantization is the effective lever. Consequently the two compression ratios differ only mildly. At $2.5$\,bpw ActQuant delivers a $5.3\times$ reduction in weight memory and a $4.0\times$ reduction in end-to-end peak VRAM ($15.3 \to 3.8$\,GB), the latter being the figure that determines whether the policy fits in the unified memory of an on-board accelerator.

\section{QVLA Reproduction and Calibration-Budget Sensitivity}
\label{app:qvla-repro}

Our QVLA~\cite{qvla} numbers in Table~\ref{tab:vla-quant-results} differ from those reported in the original paper because we run all baselines under the same low-data calibration protocol that ActQuant targets. The original QVLA paper calibrates on $512$ episodes drawn from the LIBERO fine-tuning split (out of roughly $1.7$k available episodes), which expands to over $80{,}000$ frames; computing the channel-level sensitivity scores QVLA needs requires a separate forward pass for each of those frames, at non-trivial GPU cost. ActQuant, by contrast, is designed for the extreme-PTQ regime in which calibration data is scarce, so we cap every method in our experiments at the same $60$ episodes from the LIBERO fine-tuning split (Appendix~\ref{app:calibration}). We re-ran QVLA from its public implementation under this $60$-episode budget on both OpenVLA-OFT and $\pi_{0.5}$, with all other hyperparameters left at the defaults from the original repository; the reported numbers therefore reflect QVLA's behavior at the same calibration budget as ActQuant and the other baselines, not the larger budget used in the original paper.

\paragraph{Calibration-budget sensitivity.}
Because a shared low-data protocol could in principle penalize baselines whose sensitivity statistics are more data-hungry than ours, we also quantify how much QVLA recovers when it is given its original calibration budget. We re-ran QVLA on OpenVLA-OFT with $512$ episodes ($84{,}915$ frames, roughly $30\%$ of the LIBERO fine-tuning corpus), changing only the calibration size and holding every other factor identical: the same action-space sensitivity proxy, the same $430$ target layers, the same greedy per-channel allocator over $\{2, 4, 8, 16\}$-bit, and the same evaluation harness ($4$ suites $\times$ $10$ tasks $\times$ $50$ trials $=2{,}000$ rollouts per bit-width). Table~\ref{tab:qvla-budget} reports the outcome against the $60$-episode reproduction and ActQuant, both taken from Table~\ref{tab:vla-quant-results}.

\begin{table}[h]
\centering
\caption{Effect of QVLA's calibration budget on OpenVLA-OFT, evaluated on LIBERO ($2{,}000$ rollouts per bit-width). ``Long'' denotes LIBERO-10. The $60$-episode QVLA rows and the ActQuant rows are reproduced from Table~\ref{tab:vla-quant-results}; the $512$-episode rows re-run QVLA at its original calibration budget with all other settings unchanged.}
\label{tab:qvla-budget}
\small
\begin{tabular}{@{}l l cccc c@{}}
\toprule
\multirow{2}{*}{\textbf{BPW}} & \multirow{2}{*}{\textbf{Method}} & \multicolumn{5}{c}{\textbf{Success Rate (\%)}} \\
\cmidrule(lr){3-7}
 & & \textbf{Spatial} & \textbf{Object} & \textbf{Goal} & \textbf{Long} & \textbf{Avg.} \\
\midrule
\multirow{3}{*}{4.0}
 & QVLA ($60$ ep)  & 96.0 & 98.7 & 97.4 & 95.2 & 96.8 \\
 & QVLA ($512$ ep) & 96.6 & 97.4 & 98.0 & 96.0 & 97.0 \\
 & \textbf{ActQuant} & 96.0 & 98.2 & 97.2 & 95.0 & 96.6 \\
\midrule
\multirow{3}{*}{3.5}
 & QVLA ($60$ ep)  & 82.9 & 97.2 & 91.4 & 93.6 & 91.3 \\
 & QVLA ($512$ ep) & 87.2 & 97.4 & 92.4 & 92.6 & 92.4 \\
 & \textbf{ActQuant} & \textbf{95.8} & \textbf{98.4} & \textbf{96.0} & \textbf{95.8} & \textbf{96.5} \\
\midrule
\multirow{3}{*}{3.0}
 & QVLA ($60$ ep)  & 40.8 & 56.8 & 20.2 & 31.8 & 37.4 \\
 & QVLA ($512$ ep) & 43.0 & 58.0 & 23.0 & 36.4 & 40.1 \\
 & \textbf{ActQuant} & \textbf{92.8} & \textbf{97.4} & \textbf{94.0} & \textbf{95.6} & \textbf{95.0} \\
\midrule
\multirow{3}{*}{2.5}
 & QVLA ($60$ ep)  & 0.0 & 0.2 & 0.0 & 0.0 & 0.0 \\
 & QVLA ($512$ ep) & 0.0 & 0.0 & 0.0 & 0.0 & 0.0 \\
 & \textbf{ActQuant} & \textbf{86.4} & \textbf{98.2} & \textbf{84.8} & \textbf{91.0} & \textbf{90.1} \\
\bottomrule
\end{tabular}
\end{table}

The W4A16 setting is included for completeness, as it is QVLA's own near-lossless operating point; the regime of interest here is sub-4-bit. With $8.5\times$ more calibration data, QVLA moves by at most $\sim$$3$ points at any bit-width and still degrades sharply below three bits: $40.1\%$ at $3.0$\,bpw and $0.0\%$ at $2.5$\,bpw, against ActQuant's $95.0\%$ and $90.1\%$. The degradation is uniform rather than an average masking one weak suite: at $3.0$\,bpw the per-suite scores are $43.0/58.0/23.0/36.4$, every suite between $35$ and $70$ points behind ActQuant, and at $2.5$\,bpw QVLA records zero successes across all $2{,}000$ rollouts on every suite. The cause is structural rather than statistical. The two budgets produce near-identical bit assignments, differing on fewer than $0.5\%$ of channels at every target bit-width, because below three bits the budget constraint drives QVLA's per-channel allocation toward a near-uniform $2$-bit assignment regardless of how finely the channel sensitivities are estimated; additional calibration data sharpens the sensitivity estimates but cannot relax the budget that forces this collapse. For scale, $84{,}915$ frames is roughly $660\times$ a standard LLM PTQ calibration budget ($\sim$$128$ sequences), so the sub-4-bit gap we report is not an artifact of data starvation.

\section{Closed-Loop Error Compounding}
\label{app:compounding}

The inter-tensor and intra-tensor signals of Section~\ref{sec:method} are computed open-loop, from ground-truth actions on calibration episodes, whereas the quantity we care about is closed-loop task success under a policy that visits its own states. Every success rate reported in this paper is measured closed-loop, so the open-loop proxies are already validated against the deployed objective end-to-end; Table~\ref{tab:ablation} isolates their contribution, where replacing the action-agnostic magnitude-weighted proxy with the action-guided Action-Only Fisher at a fixed bit-width gains $30.7$ closed-loop points ($57.7\% \to 88.4\%$). This appendix additionally measures the compounding mechanism itself.

We run the quantized policy closed-loop on LIBERO-Goal and query the fp16 policy at every state the quantized policy visits, recording the RMS deviation between the two action vectors as a function of rollout progress over $50$ episodes. Table~\ref{tab:compounding} compares ActQuant against RTN block quantization at $2.5$\,bpw, where the two differ sharply in success rate ($38/50$ versus $6/50$ on these rollouts).

\begin{table}[h]
\centering
\caption{RMS deviation between the quantized and fp16 action vectors, measured on states visited by the quantized policy during closed-loop LIBERO-Goal rollouts ($50$ episodes, $2.5$\,bpw). Windows are indexed by replanning step within the rollout.}
\label{tab:compounding}
\small
\begin{tabular}{@{}l c c c@{}}
\toprule
\textbf{Rollout window (replans)} & \textbf{ActQuant} & \textbf{RTN} & \textbf{RTN\,/\,ActQuant} \\
\midrule
$0$--$4$ (early)   & 0.076 & 0.086 & $1.13\times$ \\
$10$--$19$ (mid)   & 0.102 & 0.147 & $1.44\times$ \\
$30+$ (late)       & 0.116 & 0.131 & $1.12\times$ \\
\bottomrule
\end{tabular}
\end{table}

Early in a rollout both quantized policies track the fp16 reference closely and are nearly indistinguishable ($1.13\times$). The separation appears mid-rollout, exactly where compounding predicts it: RTN's deviation grows to $1.44\times$ that of ActQuant, while ActQuant's stays bounded and increases only mildly from early to late windows ($0.076 \to 0.116$). The late-window ratio narrows to $1.12\times$, but this window is conditioned on episodes that are still running, and most RTN rollouts have already failed by that point, so it reflects the surviving minority rather than a recovery. The mid-rollout separation is what the success rates record: on these rollouts ActQuant completes $38/50$ against RTN's $6/50$. The suite-dependent pattern in Table~\ref{tab:vla-quant-results} is consistent with the same mechanism, with the short-horizon Object suite remaining robust at the lowest bit-widths while the longer-horizon Goal and Long suites, which afford more replanning steps for deviations to accumulate, degrade first.

This characterizes compounding within the deployment distribution the calibration set was drawn from; how the effect changes when the deployment distribution drifts away from that calibration distribution is a separate question that we leave open.

\section{Limitations and Future Work}
\label{app:limitations}

ActQuant focuses on \emph{weight-only} quantization and leaves activations at half precision; pushing further would require pairing the per-tensor allocator with an activation-quantization scheme, which is left for future work. Our real-world evaluation serves the quantized policy from a workstation GPU rather than from a genuinely memory-constrained onboard robot computer, so it does not yet demonstrate a complete closed-loop deployment under the on-board memory budget that motivates sub-4-bit quantization. On the systems side, OmniModel.cpp inherits the operator coverage of the underlying GGML kernels, so architectures that introduce new primitives (e.g., novel attention variants or non-standard normalizations) still trigger the agent's reflector loop on first conversion; broadening the procedural-skill library to cover more architecture families is an obvious next step.

\section{Broader Impacts}
\label{app:broader-impacts}

ActQuant compresses Vision-Language-Action models so they can run on the on-board hardware of physical robots. The intended positive impact is to make capable robotic foundation models accessible without datacenter-scale GPUs, which lowers the cost of robotics research, supports applications in assistive and household settings, and reduces the energy footprint of inference because a quantized backbone draws substantially less power than its fp16 counterpart. By keeping the deployment self-contained on the robot, ActQuant also removes the privacy concerns associated with streaming raw camera data to a remote server. On the negative side, cheaper on-device VLA inference makes it easier to deploy autonomous manipulators in settings where operator oversight is weak, and a quantized policy can fail in ways that are subtly different from its fp16 reference, especially under inputs that drift from the calibration distribution; this matters because closed-loop robotic control can cause physical harm when the policy errs. We mitigate the second concern in two ways. First, our calibration and evaluation protocols are explicit about the distributions used (Appendix~\ref{app:calibration}), so deployers can match them to their target environment. Second, we never claim that the quantized model is a drop-in replacement for the fp16 model under arbitrary distribution shift, and we report real-robot success rates side-by-side with the fp16 baseline so that any drop is visible. Standard safeguards for autonomous robots, including supervised operation, geofencing, and emergency-stop hardware, remain necessary regardless of the quantization method used.

\section{LLM Usage}
\label{app:llm-usage}

We used large language models in two clearly scoped ways. Throughout the writing of this paper, we used a general-purpose assistant for sentence-level editing (clarity, concision, grammar) on prose we had already drafted. The assistant did not generate scientific claims, results, or citations; every empirical number, figure, and reference in this paper was produced and verified by the authors.

\newpage
\section*{NeurIPS Paper Checklist}

\begin{enumerate}

\item {\bf Claims}
    \item[] Question: Do the main claims made in the abstract and introduction accurately reflect the paper's contributions and scope?
    \item[] Answer: \answerYes{} % Replace by \answerYes{}, \answerNo{}, or \answerNA{}.
    \item[] Justification: the claims accurately reflect the contributions. 

\item {\bf Limitations}
    \item[] Question: Does the paper discuss the limitations of the work performed by the authors?
    \item[] Answer: \answerYes{} % Replace by \answerYes{}, \answerNo{}, or \answerNA{}.
    \item[] Justification: We dedicated a limited and future works in the appendix

\item {\bf Theory assumptions and proofs}
    \item[] Question: For each theoretical result, does the paper provide the full set of assumptions and a complete (and correct) proof?
    \item[] Answer: \answerYes{} % Replace by \answerYes{}, \answerNo{}, or \answerNA{}.
    \item[] Justification: We provided the full set of assumptions for all of our proofs. 

    \item {\bf Experimental result reproducibility}
    \item[] Question: Does the paper fully disclose all the information needed to reproduce the main experimental results of the paper to the extent that it affects the main claims and/or conclusions of the paper (regardless of whether the code and data are provided or not)?
    \item[] Answer: \answerYes{} % Replace by \answerYes{}, \answerNo{}, or \answerNA{}.
    \item[] Justification: The full instruction for result reproducibility were provided with adequate details. 

\item {\bf Open access to data and code}
    \item[] Question: Does the paper provide open access to the data and code, with sufficient instructions to faithfully reproduce the main experimental results, as described in supplemental material?
    \item[] Answer: \answerYes{} % Replace by \answerYes{}, \answerNo{}, or \answerNA{}.
    \item[] Justification: The data is publicly available and we provided access to the code for reproducibility. 

\item {\bf Experimental setting/details}
    \item[] Question: Does the paper specify all the training and test details (e.g., data splits, hyperparameters, how they were chosen, type of optimizer) necessary to understand the results?
    \item[] Answer:  \answerYes{} % Replace by , \answerNo{}, or \answerNA{}.
    \item[] Justification: All training details and implementation details are provided in the appendix.

\item {\bf Experiment statistical significance}
    \item[] Question: Does the paper report error bars suitably and correctly defined or other appropriate information about the statistical significance of the experiments?
    \item[] Answer: \answerYes{} % Replace by \answerYes{}, \answerNo{}, or \answerNA{}.
    \item[] Justification: yes the sampling details and the parameters were all mentioned in the Appendix. 

\item {\bf Experiments compute resources}
    \item[] Question: For each experiment, does the paper provide sufficient information on the computer resources (type of compute workers, memory, time of execution) needed to reproduce the experiments?
    \item[] Answer: \answerYes{} % Replace by \answerYes{}, \answerNo{}, or \answerNA{}.
    \item[] Justification: The full information about computer resources is mentioned in the Appendix. 
    
\item {\bf Code of ethics}
    \item[] Question: Does the research conducted in the paper conform, in every respect, with the NeurIPS Code of Ethics \url{https://neurips.cc/public/EthicsGuidelines}?
    \item[] Answer: \answerYes{} % Replace by \answerYes{}, \answerNo{}, or \answerNA{}.
    \item[] Justification: we follow the NeurIPS Code of Ethics.

\item {\bf Broader impacts}
    \item[] Question: Does the paper discuss both potential positive societal impacts and negative societal impacts of the work performed?
    \item[] Answer: \answerYes{} % Replace by \answerYes{}, \answerNo{}, or \answerNA{}.
    \item[] Justification: Yes. We discuss the broader impacts of ActQuant in Appendix~\ref{app:broader-impacts}, covering both potential positive impacts (lower-cost, lower-power, privacy-preserving on-device VLA inference for robotics) and potential negative impacts (easier deployment of autonomous manipulators with weaker oversight, and quantized-policy failure modes under distribution shift in closed-loop control), together with mitigations.
    
\item {\bf Safeguards}
    \item[] Question: Does the paper describe safeguards that have been put in place for responsible release of data or models that have a high risk for misuse (e.g., pre-trained language models, image generators, or scraped datasets)?
    \item[] Answer:  \answerNA{}% Replace by \answerYes{}, \answerNo{}, or \answerNA{}.
    \item[] Justification: The paper poses no such risks.

\item {\bf Licenses for existing assets}
    \item[] Question: Are the creators or original owners of assets (e.g., code, data, models), used in the paper, properly credited and are the license and terms of use explicitly mentioned and properly respected?
    \item[] Answer: \answerYes{} % Replace by \answerYes{}, \answerNo{}, or \answerNA{}.
    \item[] Justification: All codes and models are correctly credited and cited.

\item {\bf New assets}
    \item[] Question: Are new assets introduced in the paper well documented and is the documentation provided alongside the assets?
    \item[] Answer: \answerYes{}% Replace by \answerYes{}, \answerNo{}, or \answerNA{}.
    \item[] Justification: Yes. We release the ActQuant codebase and our OmniModel.cpp conversion artifacts at the anonymized link in the abstract (\url{https://anonymous.4open.science/r/ActQuant-7C06/README.md}). The repository contains a top-level README that documents installation, the calibration and quantization scripts, the bit-width sweep configurations used in Table~\ref{tab:vla-quant-results}, and the C/C++ inference build instructions. We do not release any new datasets; calibration and evaluation use the publicly available LIBERO benchmark~\cite{liu2023libero}, and the quantized model checkpoints we publish are derived from publicly released OpenVLA-OFT~\cite{oft} and $\pi_{0.5}$~\cite{pi} weights under their original licenses.

\item {\bf Crowdsourcing and research with human subjects}
    \item[] Question: For crowdsourcing experiments and research with human subjects, does the paper include the full text of instructions given to participants and screenshots, if applicable, as well as details about compensation (if any)? 
    \item[] Answer: \answerNA{}% Replace by \answerYes{}, \answerNo{}, or \answerNA{}.
    \item[] Justification: This work does not have any crowdsourcing experiments or research with human subjects.

\item {\bf Institutional review board (IRB) approvals or equivalent for research with human subjects}
    \item[] Question: Does the paper describe potential risks incurred by study participants, whether such risks were disclosed to the subjects, and whether Institutional Review Board (IRB) approvals (or an equivalent approval/review based on the requirements of your country or institution) were obtained?
    \item[] Answer: \answerNA{} % Replace by \answerYes{}, \answerNo{}, or \answerNA{}.
    \item[] Justification: This work does not have study participants

\item {\bf Declaration of LLM usage}
    \item[] Question: Does the paper describe the usage of LLMs if it is an important, original, or non-standard component of the core methods in this research? Note that if the LLM is used only for writing, editing, or formatting purposes and does \emph{not} impact the core methodology, scientific rigor, or originality of the research, declaration is not required.
    %this research? 
    \item[] Answer: \answerYes{} % Replace by \answerYes{}, \answerNo{}, or \answerNA{}.
    \item[] Justification: we dedicated a section to LLM usage in the appendix.

\end{enumerate}

\end{document}